\documentclass[times,twocolumn,final]{elsarticle}
\usepackage{medima}
\usepackage{framed,multirow}
\usepackage{subfigure}
\usepackage{amssymb}
\usepackage{latexsym}
\usepackage{makecell}
\usepackage{url}
\usepackage{xcolor}
\usepackage{diagbox}
\usepackage{hyperref}
\usepackage{subfigure}
\usepackage{caption}
\usepackage{amsmath}
\usepackage{array}
\usepackage{footmisc}
\usepackage{colortbl}
\definecolor{newcolor}{rgb}{.8,.349,.1}
\definecolor{yellow50}{RGB}{255, 255, 0}

\journal{Medical Image Analysis}

\begin{document}

\verso{Ziquan Wei \textit{et~al.}}

\begin{frontmatter}

\title{An Efficient Cervical Whole Slide Image Analysis Framework Based on Multi-scale Semantic and Location Deep Features}%
%\tnotetext[tnote1]{This is an example for title footnote coding.}

\author[1,2]{Ziquan \snm{Wei}\fnref{fn1}}
\ead{wzquan142857@hust.edu.cn}
\author[1,2]{Shenghua \snm{Cheng}\fnref{fn1}\corref{cor1}}
\cortext[cor1]{Corresponding author: Shenghua Cheng, Xiuli Liu.}
\ead{chengshen@hust.edu.cn}
\fnref{fn1}\fntext[fn1]{authors have the equal contribution.}
\author[3]{Junbo \snm{Hu}}
\ead{cqjbhu@163.com}
\author[4]{Li \snm{Chen}}
\ead{chengliisme@126.com}
\author[1,2]{Shaoqun \snm{Zeng}}
\ead{sqzeng@mail.hust.edu.cn}
\author[1,2]{Xiuli \snm{Liu}\corref{cor1}}
\ead{xlliu@mail.hust.edu.cn}

\address[1]{Collaborative Innovation Center for Biomedical Engineering, Wuhan National Laboratory for Optoelectronics-Huazhong University of Science and Technology, Wuhan, Hubei 430074, China}
\address[2]{Britton Chance Center and MOE Key Laboratory for Biomedical Photonics, School of Engineering Sciences, Huazhong University of Science and Technology, Wuhan, Hubei 430074, China}
\address[3]{Department of Pathology, Maternal and Child Hospital of Hubei Province, Tongji Medical College, Huazhong University of Science and Technology, Wuhan, Hubei 430070, China}
\address[4]{Department of Clinical Laboratory, Tongji Hospital, Tongji Medical College, Huazhong University of Science and Technology, Wuhan, Hubei 430030, China}

%\received{1 May 2013}
%\finalform{10 May 2013}
%\accepted{13 May 2013}
%\availableonline{15 May 2013}
%\communicated{S. Sarkar}

\begin{abstract}
%%%
%Digital gigapixel whole slide image (WSI) that is widely used in clinical diagnosis approximately has billions of pixels, and automated WSI analysis is key processing in computer-aided diagnosis.
%Currently, analysing the description of probabilities or feature maps from local patch encoded by classifier, such as ResNet or MobileNet, is the main manner for the WSI-level prediction.
%The spread and tiny object in cervical slides, however, is still challengeable for the under-promoted upstream encoders, and the spatial representation of cervical cells is the available feature to supply the analysis. As well as patches sampling with overlap and repetitive processing incur the inefficiency and the unpredictable side effect.
%This study designs a novel inline connection network (InCNet) by enriching the multi-scale connectivity to build the lightweight model named You Only Look Cytopathology Once (YOLCO) with the additional supervision of spatial information. The proposed model allows the input size enlarged to megapixel that can stitch the WSI without any overlap by the average repeats decreased from $10^3\sim10^4$ to $25$ for collecting features and predictions at two scales.
%Based on Transformer for classifying multi-scale multi-task features, the experimental results appear $2.12\times$ faster and $0.872$ AUC score better than the best conventional method on the WSI classification with 2,019 slides multi-cohort dataset from four scanning devices.
%%%%
Digital gigapixel whole slide image (WSI) is widely used in clinical diagnosis, and automated WSI analysis is key for computer-aided diagnosis.
Currently, analyzing the integrated descriptor of probabilities or feature maps from massive local patches encoded by ResNet classifier is the main manner for WSI-level prediction.
Feature representations of the sparse and tiny lesion cells in cervical slides, however, are still challenging, while the unused location representations are available to supply the semantics classification.
%As well as patches sampling with overlap and repetitive processing incur the inefficiency and the unpredictable side effect.
This study designs a novel and efficient framework with a new module InCNet constructed lightweight model YOLCO (You Only Look Cytology Once). It directly extracts feature inside the single cell (cluster) instead of the traditional way that from image tile with a fixed size.
The InCNet (Inline Connection Network) enriches the multi-scale connectivity without efficiency loss.
The proposal allows the input size enlarged to megapixel that can stitch the WSI by the average repeats decreased from $10^3\sim10^4$ to $10^1\sim10^2$ for collecting features and predictions at two scales.
Based on Transformer for classifying the integrated multi-scale multi-task WSI features, the experimental results appear $0.872$ AUC score better than the best conventional model on our dataset ($n$=2,019) from four scanners. The code is available in this github \href{https://github.com/Chrisa142857/You-Only-Look-Cytopathology-Once}{\textcolor{blue}{\underline{link}}}, where the deployment version has the speed $\sim$70 s/WSI.
%%%%
\end{abstract}


\end{frontmatter}

%\linenumbers

\section{Introduction}
\label{sec1}
\subsection{Background}
Cervix cancer is one of the most common and dangerous diseases with mortality, more than 0.34 million cases, which is higher than half of the incidence, about 0.6 million cases, in 2020 (world, all ages) (\cite{https://doi.org/10.3322/caac.21660}). Furthermore, the 5-year survival rate of cervical cancer is as high as $92\%$ (\cite{saslow2012american}) if detected early, and hence it is significant that enhancing the accuracy of cervical cancer screening. The usage of artificial intelligence (AI) assistance for cytopathology screening is valid and commonly based on the signal that is scanned and loaded into the computer, where the digital signal captured from the glass slide is called as whole slide image (WSI). AI assistance can directly guide pathologists starting at the recommended skeptical lesion area in the computer-aided diagnosis (CAD), instead of the traditional labour-intensive screening on the gigapixel region of interest in a real glass slide or a WSI. As such, automatically analyzing WSI and predicting at the patch-level is an active topic in the CAD.

For the automated screening, the cervical WSI with the liquid-based cytology test (LCT) is widely adopted by the medical community, which is an effective technique for aiding pathologists and computers in locating abnormal cells (\cite{davey2006effect}).
%As shown in Figure.\ref{problems} (a), based on the B system, the normal cell type named Negative for Intraepithelial Lesion for Malignancy (NILM) is uninterested for cervical screening, and the abnormal types are atypical squamous cells of undetermined significance (ASC-US), low-grade squamous intraepithelial lesion (LSIL), atypical squamous cells cannot exclude HSIL (ASC-H), high-grade squamous intraepithelial lesion (HSIL), and squamous cell carcinoma (SCC), which infection ranks from mild to severe. The WSI-level result is determined by the presentation of grades with the cytopathological experiences.
Especially, the top skeptical abnormal cells are normally considered to represent the WSI to avoid the analysis for all cells, which is the custom of human experts against the enormous count of cervical cells. Since the cytopathological diagnosis is built on the appearance and the comparison of individual abnormal cells or cells clusters under experience in the WSI along with the global information of background (\cite{nayar2017bethesda}), formulating the problem to tasks at cell level or patch-level is natural for the CAD. Unavoidably, the expert also applies the multiple dimensional features like the artificial area or the bacteria infection with the relevance between abnormal cells at the WSI-level. Consequently, researches are divided into two main directions, focusing on the local prediction only, which is the most of recent works for cervical CAD, and completing it with a WSI-level determination, which is the aim of this work to automatically analyze the LCT WSI from Papanicolaou (PAP) smears.

\subsection{Related Works}

Since the deep learning era has significantly improved the performance in various biomedical applications, the convolutional neural network (CNN) shows validity to predict cervical abnormal cells classification and segmentation at the patch-level. Many explorations based on deep learning have been studied in recent years. Emerging works are trying to take advantage of CNN to improve the segmentation accuracy of cytoplasm and nuclei (\cite{tareef2017optimizing,lu2016evaluation}). While the classification accuracy dropped with the segmentation error (\cite{zhang2017deeppap}). Then, DeepPap (\cite{zhang2017deeppap}) earlier applied CNN to classify cervical cells without the prior segmentation and hand-crafted features, which shows superior performances on the Herlev Pap smear dataset (\cite{jantzen2005pap}) and HEMLBC (H\&E stained manual LBC) dataset (\cite{zhang2014automation}), but since the random-view aggregation and multiple crop testing are parts of the inference, it is very time-consuming. To further improve the performance of classification, \cite{dong2020inception} proposed the combination of InceptionV3 and artificial features to explicitly adapt with the cervical cell domain knowledge, and achieved an accuracy of more than $98\%$ on the Herlev dataset.

Along with the development of object detecting function that able to bound and classify cervical cells, contemporarily, a customized YOLOv3 with an additional task-specific classifier head is proposed by \cite{xiang2020novel}, which achieved a mean average precision of $63.4\%$ on a 10-class dataset consists of 1,014 annotated cervical cell images with the size of $4000\times3000$. To deal with the data limitation, the customized Faster RCNN named Comparison detector is proposed by \cite{liang2021comparison} achieved an improvement against with the baseline whether on their larger dataset or their smaller dataset. Their total dataset has 48,587 instances belong to 7,410 cervical microscopical images scanned by one device. Additionally, over $16,000$ TCT (Thinprep Cytologic Test) images as the train-val set were collected in \cite{tan2021automatic}, as well as 290 larger regions of interest (ROIs) as the test set, which TCT smear is rare on cervical automatic diagnosis studies. Authors evaluated Faster RCNN enormously and randomly sampling of small images on the test set and achieved an appealing performance.

\begin{figure*}[t]
  \centering
  \includegraphics[width=\linewidth]{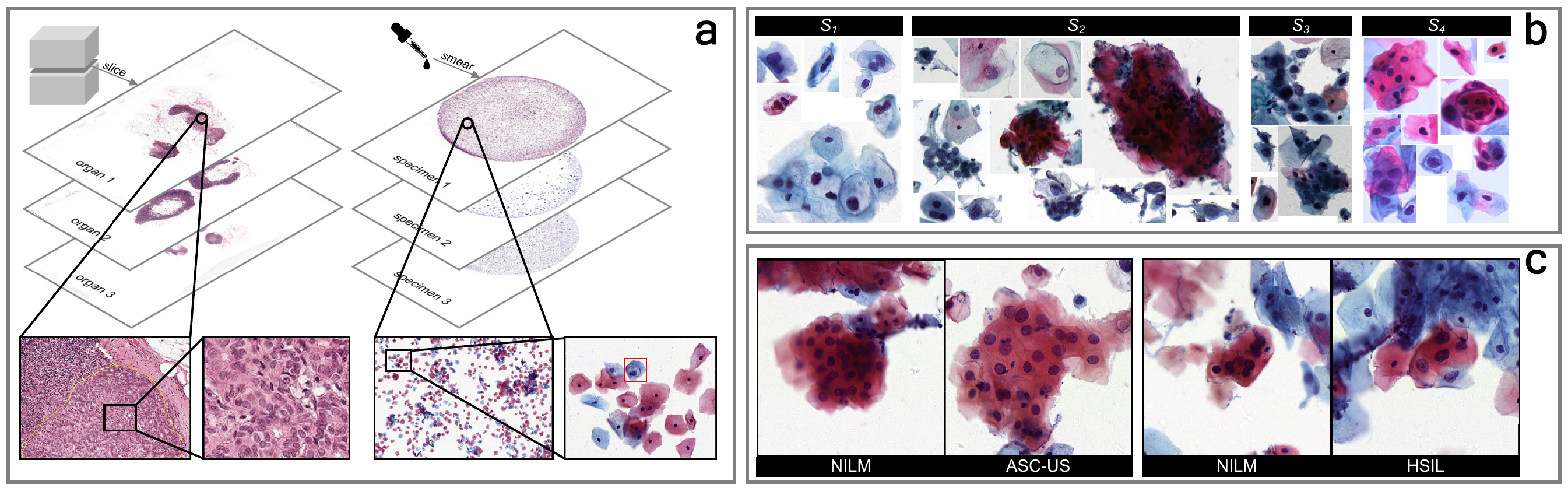}
  \caption{The example of cytopathological WSIs used in this work, which illustrates three very challenging problems of the efficient analysis of LCT WSI from PAP smears. (a) The disparity between the histopathology and the cytopathology from CAMELYON dataset and our dataset, respectively, where the lesion area is highlighted with yellow line and red box. (b) The various properties and morphologies of the Atypical Squamous Cell (ASC), where $S$ denotes different scanners, and (c) the analogous cells between normal and abnormal, which are named as Negative for Intraepithelial Lesion or Malignancy (NILM), ASC – undetermined significance (ASC-US) and High-grade Squamous Intraepithelial Lesion (HSIL).}\label{problems}
\end{figure*}

Furthermore, the cervical WSI has several gigapixels scanned at $20\times$ magnification ($0.24\mu m/pixel$), and hundreds of them can be counted of pixels equal to the entire ImageNet dataset. However, it can generate hundreds of thousands of small images that are contextual but unlabeled. For example, unlabeled foreground pixels outnumber labeled pixels of one slide 201 to 1 in our dataset. Lacking the learnable pixels is the physical property based on the sparsity as illustrated in Figure.\ref{problems} (a), unlike the nature scene is various for interesting objects and modality, or the histopathological objects are sizable and clustered. To train with the plentiful data, on the one hand, the image augmentation on color and shape transformations or results of GAN is significant on the variant cervical dataset (\cite{tan2021automatic, yu2021generative}), which is also a known trick in natural tasks and valid to deal with WSIs are multi-cohort or multicenter. On the other hand, despite there is no related work of cervical prediction, the procedure, firstly training in the primary labeled data and then complementing labels manually checked from unlabeled data predicted by trained models, is firstly proposed by \cite{bertram2019large} to efficiently augment the pathological database at the patch-level. Noteworthily, this study hypothesized that the reproducibility of pathologist-defined labels should with caution based on the improvement of this study, 0.034 $F1$-Scores, and the larger 0.186 $F1$-Scores by their further work (\cite{bertram2020pathologist}).

The most of above works aimed for the patch-level prediction is based on the ROIs pathologist preselected. However, they ignored and underestimated the automatic selection of ROIs from the entire WSI. Although the ROI identification system proposed by \cite{gupta2020region} only make a coarse prediction, their results at a rough WSI-level covered all content of cervical smears and achieved a great accuracy of tile classification on a 10 LCT WSIs dataset from PAP smears. Based on the property of histopathological WSI that objects are sizable and clustered, relative methods for the CAD are deeply developed at the WSI-level. The public CAMELYON datasets (\cite{litjens20181399}) consists of 1,399 annotated WSIs of lymph nodes to detect breast cancer also promoted the development. The grand challenge CAMELYON17 has 37 algorithms to predict at the WSI-level using 899 learnable WSIs, where the top performance from the GoogleNet was reproduced based on the multi-scale and color normalization (\cite{liu2017detecting}), which with relatively shadow architecture and relatively longer training phase than others.  After the patch-level computation with CNN, this top method used the conventional machine learning and feature engineering to classify WSIs, which was surpassed by PFA-ScanNet that using more scales to extract features (\cite{zhao2019pfa}), then it was surpassed by the novel attention-based classifier with a shadower siamese MI-FCN architecture (\cite{yao2020whole}). For the limited learnable data at the patch-level, weakly supervised learning is earlier employed for histopathological WSI-level prediction by \cite{campanella2019clinical}, authors developed the multiple instance learning (MIL)-based method on the clinical application and achieved the better performance than pathologists with a huge 44,732 WSIs dataset.

For cytopathological WSI-level prediction, to deal with thyroid WSI, \cite{dov2021weakly} proposed the modified MIL approach that the instance selection is based on maximum likelihood estimation (MLE) with additional 4,494 labeled instances achieved 0.985 of AUC on a 908 WSIs dataset. DP-Net proposed by \cite{lin2021dual} is claimed as the first deep learning-based method for the cervical WSI analysis from PAP smear. Their novel approach is a handcrafted logical classification strategy for the confusion of subclasses by automatically clustering abnormal cells in the middle hierarchy between classes, and achieved the best performance on their 19,303 cervical WSIs dataset with 560,536 patch-level annotations for the train-val phase. \cite{zhu2021hybrid} proposed an integrated system that includes 24 target detection CNNs, one patch-based classification CNN and one nucleus segmentation CNN to compute cervical LCT smears, and one XGBoost and one logic decision tree to classify at the WSI-level. They create a new 24 classes to deal with confusing subclasses for each patch-level CNN and shown an appealing result on their 81,727 smears dataset.
%To our best knowledge, there were very few works related to automatic cervical CAD at WSI-level.
\cite{cheng2021robust} proposed a robust WSI analysis for cervical cancer screening using deep learning. It is a progressive recognition method for the generalization problems of multicenter multiscanner data. They combined the low and high-resolution WSIs by two CNNs for the patch-level classification, respectively, and enriched the augmentation of various train data. The WSI-level probabilities are computed by another recurrent neural network (RNN) based on the semantic feature from the high-resolution CNN. Their proposal achieved an appealing performance on their 3,545 WSIs based on 79,218 patch-level annotations.

The currently automatic cervical WSI analysis works, however, are not only based on the massive cost of the patch-level annotations but the WSI-level classification is not implemented by the affluent informational features. They generally established with a large number of manual annotations and gathered the patch classification probabilities or the single semantic features by a single classification task to represent the WSI. In this work, WSIs of our dataset are represented as the semantic and location features by dual tasks, that is classification task + location task (a.k.a. object detection).

In summary, there are three challenging problems of the efficient analysis of LCT WSI from PAP smears:
\begin{enumerate}[i)]
  \item For the overall WSI analysis, as illustrated in Figure.\ref{problems} (a), the physical properties of cytopathology are different with histopathology. The former lesion zones are relatively few and small, while the latter is ubiquitous and sizable in lesion zones. Frameworks used in histopathology would be disadvantageous if disregarding such sparsity and subtlety of valuable objects in cytopathology WSIs.
  %\item For the local lesion area prediction and representation, the variant grades of infection cause the flexible complexity, e.g., the most confusing atypical squamous cells (ASC), based on the Bethesda system (\cite{nayar2015bethesda}). As well as the morphological texture varies, and its discrepancy between abnormal and hard normal cells is slight. Those cause the analysis is highly dependent on the experience of cytopathologists. Several examples are shown in Figure.\ref{problems} (b) and (c). % in the shape of the nucleus and the context of cytoplasm % and  %  are various of all parts of the cell, the nucleus shape, the cytoplasmic texture, and nucleus to cytoplasmic ratio
  \item Unlike natural tasks, pixels of pathology image are huge but the variety of texture is relatively limited. Although it was proved that deeper model can lead more performance by extracting the deep feature, cervical data needs more efficiency than depth of the model to complete the WSI-level task. As well as the discrepancy between abnormal and hard normal cells is slight, which causes the local lesion detection is highly dependent on the experience of cytopathologists. Several examples are shown in Figure.\ref{problems} (b) and (c).
  %\item The huge-size pixels are inevitably needed to be computed while the annotation is high-cost, and hence the efficiency of CNN poses the further challenge to automatically analyze.
  \item Traditional frameworks for WSI-level task adopted the overlap sampling method that repeatedly computes image tile to reduce the severe cell truncation, which is a side effect that cells located in the edge of image could be cut. The performance will be thus reduced when meeting truncated cells. The redundant computation of overlap pixels, however, further drops the efficiency of WSI-level task. %non-existed in the train phase.
\end{enumerate}

The challenges and the problems mentioned above of the cytopathological tasks pose the specific design of CNN in this work. Firstly, feature representation and feature variety are the main factors of the CNN encoder are considered. Secondly, the model that able to provide the summary and brief feature should be lightweight, for workload reduction. Thirdly, the performance of the WSI-level classification should be not only great for WSIs from a single source but stable for different scanners and cohorts. % possess the patch-level annotation

\subsection{Our contributions}

The shape annotation experts draw to bound the abnormal cells at the patch-level is credibly reproducible caused by non-cytopathological experiences. While the cytopathological experiences on the cervical content are dissimilar between different experts. Such noise of annotations can affect the performance of CNN, the feature representation for instance, especially for multi-cohort or multiscanner WSIs. The natural detection network includes feature maps fused the local size information and the object content into a single model that accomplishes dual tasks of lesion locating and content classifying, such as YOLO series (\cite{redmon2017yolo9000, redmon2018yolov3}). Their potentiality to extract semantic and size features inspires us. We design an efficient cervical WSI analysis framework based on multi-scale semantic and location features using deep learning, and contributions are three points as follows corresponding the challenges:
\begin{enumerate}[i)]
%  \item Based on the multi-task deep learning of YOLO series, proposing a lightweight CNN named YOLCO to represent cervical WSI, which is much lighter than common models, with the simplified computational method for the efficiency, and the mixed multi-scale semantic and spatial features for sparse and tiny cells.
%  \item The novel inline connection network (InCNet) proposed enriches the connectivity of various size receptive fields inner the module to enhance the representation of cervical features with a lighter structure than conventional networks.
%  \item We enlarge the input size to megapixel for multi-scale multi-task features collecting and predicting at two scales based on Transformer. The significance of our method is proved on 2,019 slides dataset with multiple familiarities of data and multiple scanners, where performances are superior comparing competitive methods.
    \item For the challenge i), we choose to design a detection model named YOLCO (You Only Look Cytology Once) to build our framework. Its feature mixes semantic and location information. Simultaneously, the feature is directly extracted inside the single cell/cluster instead of the traditional way that from image tile with a fixed size. According to this, we can avoid the cervical feature remaining at the fixed tile and the only semantic level.
    \item For the challenge ii), we choose to enhance the performance of lightweight network. We proposed an inline connection network (InCNet) to replace normal convolution network. It has several groupwise connection inside one layer, and constructed by the efficient depthwise separetable (DS) convolution. The proposal can improve the performance of lightweight model without increasing depthes and parameters for cervical data.
    \item For the challenge iii), we choose to extremely increase the size of input tile in our framework. Since the proposed YOLCO has the lightest weight, we can increase one side length of image tile from $10^2$ to $10^4$ level pixels, which can be directly adopted into our framework. With such megapixel input, the proposed method can compute one WSI in only 43 repeats, averagely, and it needs no overlap sampling cause that above side effect is weak in the small number of input.
\end{enumerate}

\section{Materials}
\label{sec:materials}

As shown in the table \ref{datasets}, nine cohorts of WSIs from the lab of the hospital\footnote{\label{hospital}The Maternal and Child Hospital of Hubei Province} are used in this work by four different scanners: $S_1$ - the version 1 of 3DHisTech Ltd., $S_2$ - the version 2 of 3DHisTech Ltd., $S_3$ - Shenzhen Shengqiang Technology Ltd., $S_4$ - Wuhan National Laboratory for Optoelectronics-Huazhong University of Science and Technology. Each device is set with a different parameter for scanning. All WSIs are sampled under $20\times$ magnification with a subtle change of resolution except $S_3$ under $40\times$. Version 1 and version 2 of 3DHisTech Ltd. are different in slide preparation and staining scheme. Furthermore, each WSI includes average $4.5$ billion pixels at the central circular foreground using OTSU method (\cite{otsu1979threshold}).

Total 19,974 typical lesion areas annotated of 138 WSIs from the $1^{st}$ and $2^{nd}$ cohorts which overall positive cervical cells, atypical squamous cells of undetermined significance (ASC-US), low-grade squamous intraepithelial lesion (LSIL), atypical squamous cells cannot exclude HSIL (ASC-H), high-grade squamous intraepithelial lesion (HSIL), and squamous cell carcinoma (SCC), are grouped by experts in one foreground class in this work. For WSI-level annotations, the $8^{th}$ cohort has manually labeled with an overall subclass by experts, and there are 39 ASC-US, 27 LSIL, 25 HSIL. While others have only binary annotations.

\begin{table}[!t]\footnotesize
\caption{Nine cohorts of WSIs from the lab of the hospital\protect\textsuperscript{\ref{hospital}} by four different scanners are used in this work, where $N_{ann.}$ denotes the number of bounding box annotations, $neg.$ and $pos.$ denote the normal slides and the abnormal slides, respectively.}\label{datasets}
\centering
%\begin{tabular}{p{0.9cm}p{1.4cm}lp{0.8cm}p{0.7cm}p{0.8cm}}
\begin{tabular}{lllllll}
\hline
scanner & cohort & res$_{\mu m/pixel}$ & slides & $pos.$ & $neg.$ & $N_{ann.}$ \\\hline
$S_1$ & $1^{st}$ & 0.24 $(20\times)$ & 208 & 103 & 105 & 16,914 \\\hline
\multirow{2}*{$S_2$} & $2^{nd}$  & \multirow{2}*{0.24 $(20\times)$} & 537 & 160 & 377 & 3,060\\
 & $3^{rd}$ &  & 221 & 120 & 101 & 0\\\hline
\multirow{3}*{$S_3$}  & $3^{rd},5^{th}$ & \multirow{3}*{0.18 $(40\times)$} & 166 & 166 & 0 & 0 \\
 & $6^{th},7^{th},8^{th}$ &  & 512 & 194 & 308 & 0 \\
 & $9^{th}$ &  & 66 & 66 & 0 & 0 \\\hline
$S_4$& $3^{rd},4^{th},5^{th}$ & 0.29 $(20\times)$ & 319 & 124 & 195 & 0 \\\hline
%xyw12 & $L_2$ & SZSQ & 0.180 $(40\times)$ & 79 & 47/32 & null \\
Total & & & 2,019 & 933  & 1,086 & 19,974\\\hline
\end{tabular}
\end{table}

\begin{figure*}[ht]
  \centering
  \includegraphics[width=\linewidth]{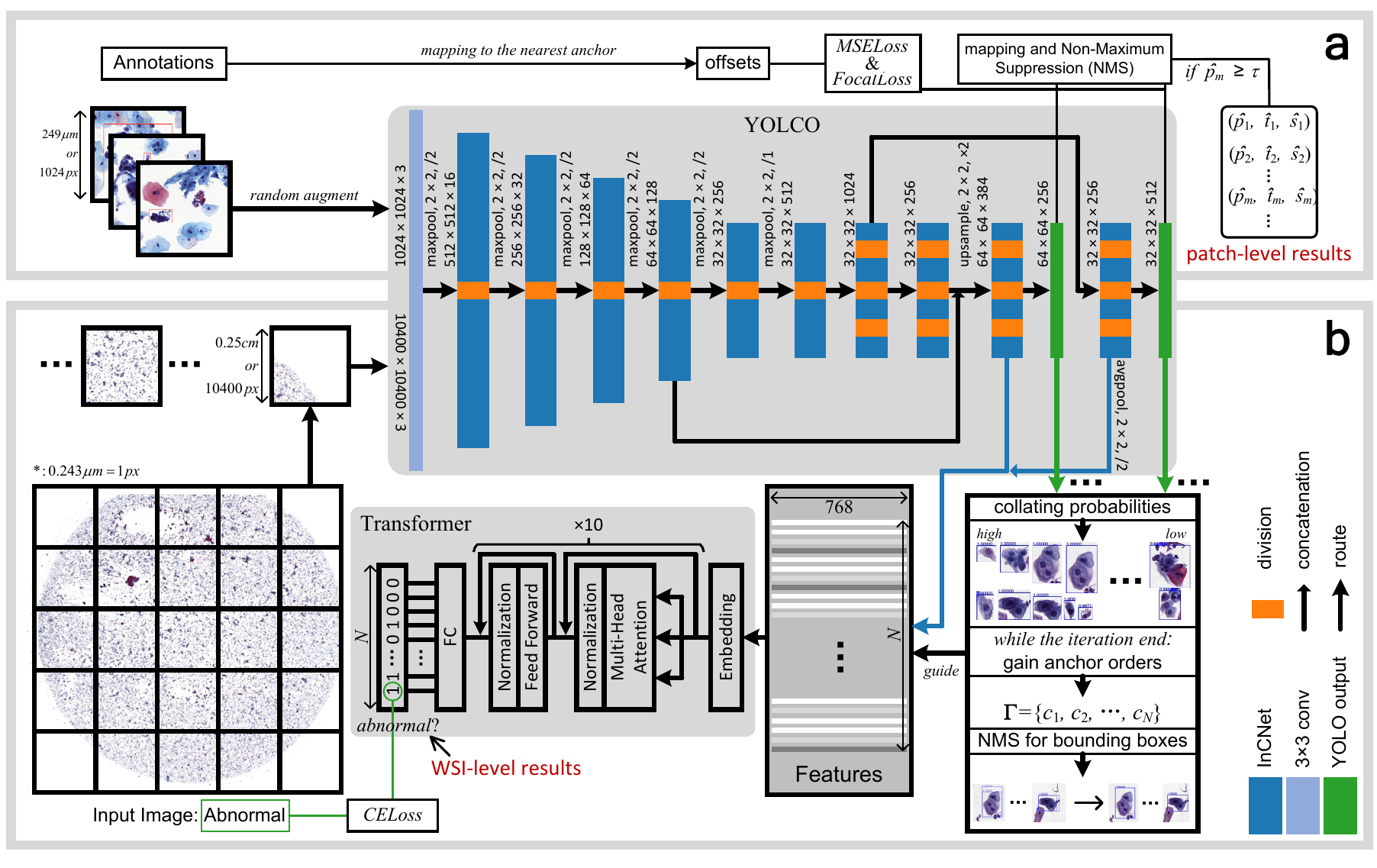}
  \caption{The pipeline of the proposed method. (a) The process for cervical lesion detection at the patch-level in preselected ROIs. (b) The process for the automatic cervical diagnosis for the normal/abnormal classification at the WSI-level with top-$N$ leison area.}\label{pipeline}
\end{figure*}

\section{Method}
\label{sec:method}

This work proposes the simplistic pipeline of the efficient cervical WSI analysis via computing much bigger ROI in few repeats of the single model with rethought of the connection approach in CNN. Increasing the input size from $10^2$ to $10^4$ is credibly to decrease the side effect, that is the truncated cells when cropping small images, and the additional processing, since WSIs are normally $10^4$ pixel level of one side in our datasets.
%The multiple scale feature extraction is employed with multiple task learning, which is an efficient trick to the object recognition by the single model.

For merging the mixed features in the cytopathological WSI computation and recognizing the cervical cells, the design is described in this section. Furthermore, the complex feature of cytopathology is enhanced by the novel InCNet as well as its principle and interpretation are depicted.

\subsection{Framework}
\label{sec:main-process}

To start with, briefly describing the pipeline of our method to analyze WSIs that consists of two main parts. As shown in Figure.\ref{pipeline} (a), firstly, training the lightweight model YOLCO at the patch-level, which has the same structure of the tiny-YOLOv3 (Tiny) and is built by lighter InCNet modules with the depthwise separable (DS) convolution, to attain the WSI encoder and the abnormal cells detector at the same time. The function that enhances the cervical representation provided by InCNet (see Section.\ref{sec:icn}) is plugged throughout the model except for the input layer and the output layer of each scale level. LeakyRELU is the activation for each layer except the output layer. YOLCO consists of two levels of the scale with $2^5$ the downsample ratio at $9^{th}$ conv-layer or $2^4$ the downsample ratio at $10^{th}$ conv-layer in two branches, respectively. Another part, Figure.\ref{pipeline} (b), is the efficient diagnosis achieved via the property of YOLCO of changeable input size and super lightweight to strikingly reduce the repeated computations, based on the same magnitude of each side between stitched patches and the central forground in the WSI. YOLCO improves the speed of encoding and recognizes the WSI as the common detection task. The sequence of features mixed at two levels of the scale from the input of every output layer is collected via top-$N$ probabilities guided of the corresponding non-repetitive coordinate of the anchor before the non-maximum suppression (NMS), which would expel neighboring feature vectors that cooperatively represent complex objects. The WSI-level classification is completed with an additional sequence classifier, Transformer (\cite{vaswani2017attention}) or simpler traditional methods (see Section.\ref{sec:exp-classifier}), for the WSI feature embedding.

For the details of training YOLCO, the patch-level labeled data is randomly cropped as $1024\times1024$ RGB images enable to embrace every bounding box in our datasets. Assuming $M$ annotations for cervical lesion is denoted by $\mathbf{T}= \{\mathbf{t}_m\},\mathbf{t}_m\in\mathbb{R}^2$ the center point of lesion area, $\mathbf{S}=\{\mathbf{s}_m\},\mathbf{s}_m\in\mathbb{R}^2$ the size of lesion area and $\mathbf{P}=\{{p}_m\},{p}_m\in\{0, 1\}$ the target label for determining positive or negative. For the prediction and WSI representation at the patch-level, anchors are the common ingredients to train the detection CNN, which transmute the problems from solving Euclidean coordinates and sizes of boxes to estimating offsets of the nearest anchor denoted $\mathbf{G}=\{\mathbf{g}_c\}, \mathbf G\in\mathbb{N}^{2\times C(l)}$ the coordinate of grid points, where $C(l) = w(l)\times h(l)$ the width times the height of the $l^{th}$ layer, and $\mathbf{A}=\{\mathbf{a}_i\}, \mathbf{A}\in\mathbb{R}^{2\times n_a}$ the size of the anchor with $n_a$ the number of shapes. Three sizes of the anchor are set via the $k$-means clustering with $k=3$ in this work. Then the annotation is mapped as $\mathbf{B}=\{\mathbf{q}_m, \mathbf{v}_m\},\mathbf{B}\in\mathbb{R}^{(2+2)\times M}$ the offset of anchors to supervise the prediction of YOLCO, where
\begin{equation}
\begin{aligned}\label{eq0}
  \mathbf{q}_m &= \mathop{\arg\min}\limits_{\mathbf{t}_m - \mathbf{g}_c}\|\mathbf{t}_m - \mathbf{G}\|^2, \\
  \mathbf{v}_m &= \mathop{\arg\max}\limits_{log({\mathbf{s}_m}/{\mathbf{a}_i})}iou_{s}(\mathbf{s}_m,\mathbf{A}),
\end{aligned}
\end{equation}
and $iou_{s}$ is the ratio of the inter area to the union area between two shapes of the box with the same coordinate. The order of the nearest anchors, $\mathbf O = \{\mathbf o_m\}=\{\mathop{\arg\min}_{c}\|\mathbf{t}_m - \mathbf{G}\|^2, \mathop{\arg\min}_{i}iou_{s}(\mathbf{s}_m,\mathbf{A})\}, \mathbf O\in\mathbb{N}^{2\times K}$, is collected in the meantime.

Consequently, the output of YOLCO is fitted as the offset of every anchors that named as $\mathbf{Y}^{box}=\{\mathbf{\hat{q}}_{c,i}, \mathbf{\hat{v}}_{c,i}\}, \mathbf{Y}^{box}\in\mathbb{R}^{(2+2)\times C(l)\times n_a}$ and $\mathbf{Y}^{cls}=\{\hat{p}_{c,i}\},\mathbf{Y}^{cls}\in\mathbb{R}^{C(l)\times n_a}$, for each anchor of each grip point in the feature map, then we employ the mixed loss function as follow:
\begin{equation}\label{eq1}
  \mathcal L_{total} = \mathcal L_{box}(\mathbf B, \mathbf Y_{obj}^{box}) + \alpha\cdot\mathcal L_{cls}(\mathbf P, \mathbf Y_{obj}^{cls}) + \beta\cdot\mathcal L_{cls}(\mathbf P, \mathbf Y_{noobj}^{cls}),
\end{equation}
where $\mathbf Y_{obj}^{\circ}=\{\mathbf Y^{\circ}|\{c,i\}\in\mathbf O\}, \mathbf Y_{noobj}^{\circ}=\{\mathbf Y^{\circ}|\{c,i\}\notin\mathbf O\}$, and $\alpha=1, \beta=100$ control the balance between the coordinates and sizes of object and the classification of foreground and background.
\begin{equation}
\begin{aligned}\label{eq2}
  \mathcal L_{box} &= MSELoss(\mathbf{B}, \mathbf{Y}^{box}) \\&= \frac{1}{K}\sum_{k}^{K}(\|\mathbf q_m-\mathbf{\hat{q}}_m\|^2+\|\mathbf v_m-\mathbf{\hat{v}}_m\|^2) ,
\end{aligned}
\end{equation}
\begin{equation}
\begin{aligned}\label{eq3}
  \mathcal L_{cls}=& FocalLoss(\mathbf{P}, \mathbf{Y}^{cls}) \\=& -\frac{\theta}{K}\sum_{k}^{K}(1-\hat{p}_m)^\gamma p_m log({\hat{p}_m}+\epsilon) \\
  &-\frac{1-\theta}{K}\sum_{k}^{K}(\hat{p}_m)^\gamma (1-p_m) log(1-{\hat{p}_m}+\epsilon),
\end{aligned}
\end{equation}
where $\theta=0.95, \gamma=1, \epsilon=1e-9$ are parameters of the focal loss with default settings (\cite{lin2017focal}). Naturally, the predicted offset of anchors will be gathered by $\tau$ the probability threshold in the patch-level test phase without $\mathbf O$, then $\{\mathbf{\hat T}, \mathbf{\hat S}, \mathbf{\hat P}\}=\{\mathbf{\hat{q}}_{c,i}+\mathbf g_c, exp(\mathbf{\hat{v}}_{c,i})\cdot\mathbf a_i,\hat{t}_{c,i}|\hat{t}_{c,i}\geq\tau\}$ the predicted boxes by inversely operated from the Equation.\ref{eq0} are transported to NMS for the patch-level results.

\begin{figure}[!t]
  \centering
  \includegraphics[width=.9\linewidth]{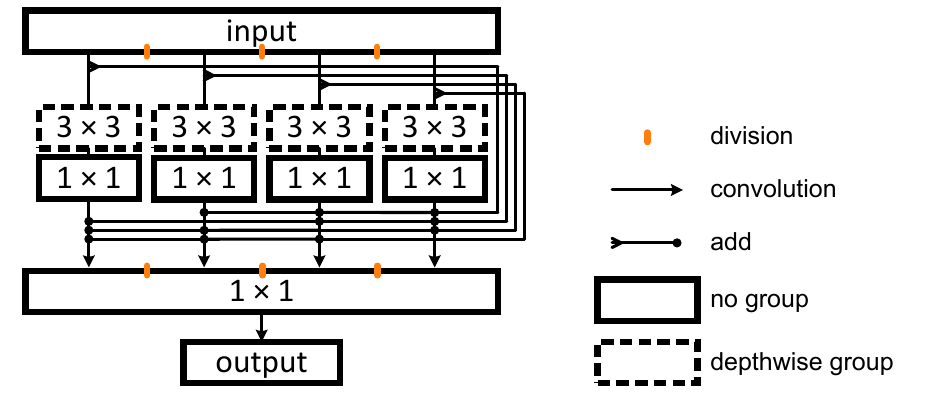} %ICN.pdf
  \caption{The proposed convolutional inline connection network (InCNet) with 3 times of division for example.}\label{icn} % The revolution of the skip connection networks, and the proposed inline connection network (InCNet).
\end{figure}

\begin{table*}[!t]
  \centering
  \caption{The implement and properties of CNN encoders in experiments, where $N_c$ is the total length of the feature sequence from $\eta$ scales, or the single scale with $\eta=1$, $N_{tile}$ the average number of input tiles from WSI, $CE$ the cross entropy loss, and MACs the Memory Access Costs, which means multi-add calculation, calculated by the python tool, thop. }\label{implement} % YOLOv5s and YOLOv5l by  \cite{glennjocher20214679653} the newest YOLO model, % LR is the learning rate and BS is the batch size MNV2 denotes the MobileNetV2 (\cite{sandler2018mobilenetv2}) and Tiny denotes the YOLOv3tiny. YOLOv3 and Tiny by \cite{redmon2018yolov3}, and InCNet$_{n_g=8}$, YOCO and YOLCO proposed in this work are implemented under PyTorch.
  \begin{tabular}{p{1.25cm}p{1.6cm}p{1.6cm}p{1.5cm}p{1.4cm}p{1.85cm}lp{0.6cm}ll}
    \hline
    % after \\: \hline or \cline{col1-col2} \cline{col3-col4} ... %\multirow{2}*{CNN}
      & \multicolumn{2}{c}{Input side length $(px)$} & \multirow{2}*{Params$(M)$} & \multirow{2}*{$\mathop{\text{MACs}(G)}\limits_{1024\times1024\times3}$} & \multirow{2}*{Loss func.} & \multirow{2}*{Data augments} & \multirow{2}*{$N_{tile}$} &  \multirow{2}*{$N_c,\langle\eta\rangle$} & \\\cline{2-3}%\multirow{2}*{$\mathop{\text{Speed}}\limits_{s/\text{WSI}}$}\\\cline{2-3}% & \multirow{2}*{LR} & \multirow{2}*{BS} \\\cline{2-3} \multirow{2}*{\% of FG} &
     & Patch-level & WSI-level \\\hline
    MNV2 & \multirow{2}*{320} & 8320 & $2.23$ & $6.54$ & \multirow{2}*{$CE$} & \multirow{2}*{shift, normalize} & 68 & \multirow{2}*{1280$,\langle1\rangle$} & \\%848.1\\%& 0.0001 & 64 \\\hline
    InCNet$_{n_g=8}$ & & 6944 & $2.79$ & $10.08$ & & & 98 & & \\\hline%896.3\\\hline%& 0.0001 & 64 \\\hline
%    YOLOv5s & 1024 & - & $7.3M$ & - & - & CIoU, CE & \multirow{2}*{yolov5 augments} & \multirow{2}*{0.00265} & \multirow{2}*{4} \\
%    YOLOv5l & 1024 & - & $47.8M$ & - & - & CIoU, CE & & &  \\\hline
    AFR & 1024 & 2080 & $41.98$ & $148.85$ & $MAE$, $CE$ & \makecell[l]{flip, rotate, scale, \\blur} & 1276 & - & \\%1097.4 \\%& \multirow{4}*{0.00001} &
    YOLOv3 & 1024 & 2080 & $61.52$ & $198.47$ & $MSE$, $Focal$ & \makecell[l]{shift, scale} & 1276 & 1792$,\langle3\rangle$ & \\\hline%1352.6 \\\hline%& \multirow{4}*{0.00001} & \multirow{4}*{4} \\
    Tiny & \multirow{3}*{1024} & \multirow{3}*{10400} & $8.67$ & $16.60$ & $MSE$, $Focal$ & \multirow{3}*{\makecell[l]{shift, scale}} & \multirow{3}*{43} & \multirow{3}*{768$,\langle2\rangle$} & \\%572.6 \\%& & \\
    YOCO &  &  & $1.79$ & $3.55$ & $Focal$ & & & & \\%538.0 \\%& & \\
    YOLCO &  &  & $1.79$ & $3.55$ & $MSE$, $Focal$ & & & & \\%538.0 \\%& & \\
    \hline
  \end{tabular}
\end{table*}

For the details of the WSI-level analysis as shown in Figure.\ref{pipeline} (b), the RGB WSI is split with size of $10400\times10400$ as the same magnification as the patch-level from the center to the peripheral of the circle of the smear using OTSU (\cite{otsu1979threshold}) at scale-level 8, where scale-level means the image is downsampled $x$ times using $ratio=2$ each time ($x=8$ in this case). Comparing with the patch-level, the threshold is set as $\tau=0$ for top-$N$ collecting by the order $\mathbf\Gamma = \mathop{\arg\text{sort}}_{c}\mathbf{\hat P}$, $\mathbf\Gamma\in\mathbb{N}^{2\times C(l)\times n_a}$, where the collecting approach is considered with the spatial restriction (see Section.\ref{sec:collecting}). Then mapping $\mathbf X$, $\mathbf X\in\mathbb{R}^{768\times N}$ the set of lesion area feature embeddings in a given WSI to $\mathbb{R}^{D_{FF}\times N}$ as $\mathbf{\hat X}$ by Transformer with $D_{FF}$ the dimension of Feed Forward layer. Finally, the WSI-level results are attained via $\text{NMS}(\mathbf{\hat T}_{\mathbf\Gamma},\mathbf{\hat S_{\mathbf\Gamma}})$ to detect abnormal cells, and the last prediction of $sigmoid(\mathbf W_{mlp}\mathbf{\hat X})$, with the parameter of the last full connection (FC) layer, $\mathbf W_{mlp}\in\mathbb{R}^{n_{cls}\times D_{FF}}$, is supervised by the cross entropy loss ($CELoss$), where the number of class $n_{cls}=2$.

\subsection{Inline Connection Network (InCNet)}
\label{sec:icn}

Since the complexity of cervical objects varies, it is left to task heads, FC layer, or the pointwise convolution layer of natural CNNs whether the multistep (\cite{ren2015faster}) or the single step (\cite{redmon2018yolov3}), which are burdened with additional problems in this special scene while the task supervision is the most attentive.

As shown in Figure.\ref{icn}, the InCNet is inspired by the group convolution and has a similar motivation with the inner skip connection (Res2Net by \cite{ISI:000607383300018}). It poses the multiple groupwise connections inner the module to enrich the connectivity between $3\times3$ filters and $1\times1$ filters with a lighter structure, which is based on the concept that the DS convolution acts the same function of the normal convolution (\cite{howard2017mobilenets}). The receptive field (RF) for the cervical cell image hardly encodes all information with the single size filter, since the local pattern of the sparse lesion is more important and complex than the dense region that covers the normal cell or the background. The groupwise connection between them guides multi-scale RFs and enhances the performance of the local pattern. Especially, the connection attaches the output of $3\times3$ filters for a bigger RF and $1\times1$ filters for a smaller RF of $l^{th}$ layer as the same level of nonlinearity:
\begin{equation}\label{eq4}
\mathbf x_{l+1} = \text{LeakyRELU}(\mathcal P(\{\mathcal D(\mathbf x_l[\iota]) + \sum^{j\neq\iota}\mathbf x_l[j]\}_{\iota=1\sim n_g}))
\end{equation}
where $\mathcal P, \mathcal D$ are the pointwise convolution and the DS convolution, respectively, $x_l[\iota]$ is the $\iota^{th}$ group of the input of $l^{th}$ layer with $n_g$ is the count of group with $n_g-1$ times of even division in the channel direction.

Moreover, the skip connection is non-considered for the efficiency of the proposed model caused that the shallow CNN is viable to propagate gradients of each node even the deepest node without the residual learning (\cite{he2016deep}).

%\subsection{Learning to recognize cervical cells}
%
%The full convolution structure of YOLCO allows a flexible input size. The saved free memory from gradients in the training stage that could be allocated for a bigger input in the inference stage is more efficient than the fixed size strategy based on a frozen computation graph. With flexible ROI size.
%
%\subsubsection{zero-padding problems}
%
%The influence of zero-padding problem illustrates a decrease along enlarging of the input size and reducing of the CNN depth. Cutting the number of layers to reduce the receptive field of the model. Enhancing the invariant feature extraction with multiple size input.

\section{Train and Inference}

As mentioned in the method section, for the WSI classification, models of the proposed framework need first to train at patch-level, and then to compute WSIs. The final WSI classification result is depends on the representation of cells at patch-level. Meanwhile, the abnormal cell detection performance can show which areas lead the final result. Thus, in this section, we describe the procedure of train and inference of every model compared in our experiments, and the tarin-val-test data setting at both patch-level and WSI-level.

\subsection{Models in experiments}

Total 7 CNNs as shown in Table.\ref{implement} are trained with annotated bounding boxes among $1^{st}$ and $2^{nd}$ cohorts in our experiments.
The first consideration for choosing compared models is the framework:
\begin{itemize}
  \item The related work for WSI analysis by \cite{CAO2021102197} consists of a detector, AttFPN-Faster-RCNN (called AFR), and a patch image classifier, ResNet50. Its framework is similar with the proposal and has three steps of WSI inference: (i) detecting bounding boxes, (ii) classifying the input image of detector by the detected top-4 boxes, and (iii) predicting the WSI classification using the mean probability of top-10 images. Thus, we use the same setting as our framework to train AFR + ResNet50.
  \item The proposed framework has two steps of inference: (i) detecting bounding boxes that separately correspond to one feature vector, (ii) classifying the collection of feature vectors of top-$N$ boxes as the WSI-level class. The proposed YOLCO is the detector in this framework. It is based on the version 3 of YOLO by \cite{redmon2018yolov3} (called YOLOv3) and its Tiny version (called Tiny). YOLOv3 and YOLCO have same structure but different backbone. The difference between Tiny and YOLCO is only the replacement of normal convolutional layers by InCNet layers using the same channel number of input and output, and they have more balanced parameters.
  \item To generate feature vector, traditional CNN encoder is supervised by the classification task instead of detecting bounding box. Therefore, we test an imperfect version of the proposed framework by using classifier to replace the detector, including the state-of-the-art (SOTA) MobileNetV2 by \cite{sandler2018mobilenetv2} (called MNV2), InCNet$_{n_g=8}$, and YOCO. The difference between MNV2 and InCNet$_{n_g=8}$ is only the replacement of normal DS layers by InCNet layers using the same channel number of input and output. YOCO is the strict contrast version, which has the same architecture and training process as YOLCO but the removal of the box representation ($\mathcal L_{box}$), to display the significance of dual tasks.
      % Different settings of grouping $n_g$ are also test at patch-level (see Section.\ref{sec:ablation-patch-level}), and the model profile is listed with the best $n_g=8$.
\end{itemize}

Aside from the previous work AFR that is strictly tested as same as the public code, according to the weight and the supervision of CNNs, different setting of $O$ and train-val-test procedure are used in our framework, respectively.

First, the overlap of input image sampling in the inference is set as $O=0$ for the lightweight CNNs because the input size can be extremely increased, including MNV2, InCNet$_{n_g=8}$, YOCO, and YOLCO, while YOLOv3 is set as $O=288$. For $O\neq 0$, the post-processing, which stitches feature maps in the overlap area, is a max sampling approach with the probability maximum.

Second, the data augmentation in Table.\ref{implement} is randomly applied in the training phase of both detectors and classifiers with $p=0.5$. There are 200 epochs at most of the learning at patch-level.

For training detectors, the input image is size of $1024\times 1024$ pixels, and output feature vectors is size of $32\times 32\times N_{ch}$ with the channel number $N_{ch}$. While the input is $320\times 320$ for training classifiers, like MNV2, and the output is $1\times 1\times N_{ch}$.

Thus, for the inference, except for YOCO and other detectors that output feature vectors corresponding the area is size of $32\times32$ pixels, feature vectors of MNV2 and InCNet$_{n_g=8}$ are $320\times320$. The enlarged input size at WSI-level of inference as listed in Table.\ref{implement} is set based on the storage capacity of the GPU.

Finally, feature sequence classifiers, e.g. Transformers, are all learning for 300 epochs. All trains have the same learning schedule and beginning at $lr=0.000005$.

Furthermore, all the classification experiments at both patch- and WSI-level consist of six loops of train-val-test because of limited number of our slides.

%\begin{figure}[!t]
%  \centering
%  \includegraphics[width=.8\linewidth]{explain_disthres.pdf}
%  \caption{The influence of different collection setting of feature vectors.}\label{collecting_method}
%\end{figure}

\subsubsection{Sequence classifiers in our framework}

For certifying the ability of feature representation of the proposed method, four types of classifiers are discussed:

(i) SVM, the conventional machine learning method using the default setting implemented by the scikit-learn toolbox, (ii) RNN, the recurrent neural network using one recurrent layer and 2048 features in the hidden state and being implemented by PyTorch,

(iii) LSTM, the recurrent neural network with memory function as the same parameters setting with RNN,

(iv) Transformer\footnote{Transformer codes are from \url{https://github.com/lucidrains/vit-pytorch}}, the neural network solely based on attention mechanisms using the recommended parameters setting that repeats Transformer layer 10 times and the number of hidden features is 2048.

Except for SVM, all classifiers are training with a dropout layer with a 0.5 drop rate, considering the avoid overfitting for their FC layers. Note that the sorting order of the count of parameters is (iv)$>$(iii)$>$(ii), and (i) is directly computing the hyperplane of all vectors of the input sequence flatten into one column without additional parameters.

\subsubsection{WSI-level feature Collecting in our framework}
\label{sec:collecting}

The collection of cervical feature extraction is the smallest step but not least.
%Bounding boxes of potential lesion area and the location information are influence factors of performance of our framework.
The reason is that the feature vector from detectors is corresponding to a smaller area (in our case, $32\times32$) than the classifiers (e.g. $320\times320$ in MNV2). Therefore, since the size of abnormal cell is larger than the computing area, the number of vectors will be various if there is no limitation of collection. Two collecting methods are designed to assess:

(i) The normal one is a Euclidean distance threshold $D$ to restrict feature vectors of top-$N$ boxes, where $N\leq500$ considering that the long sequence classification is challenging. The collecting area with lower score will be reject if the distance between two areas is smaller than $D$. %, and $D\geq0$ is working as Figure \ref{collecting_method}

(ii) \textsl{Bbox} is the other one that guided by the predicted bounding box of the lesion area via collecting $10$ feature vectors inside each top-$N$ boxes, which means $10\times N$ vectors will be gathered.
% In addition, unlike no needing for the extra computation of different $N$, each different parameter $D$ or \textsl{Bbox} needs to enroll a new computation, thus, the total mixed data are used when controlling $N$ and fixed $D$ or \textsl{Bbox}, and the smaller seen data are used when controlling $D$ and \textsl{Bbox}.

\subsection{Data Setting}

\subsubsection{Private data}
\label{sec:4.2}

%\begin{table}[t]
%\caption{Setting of train and validation in the detection task, where bbox denotes the bounding box.}
%\label{detdata-split}
%\setlength{\tabcolsep}{4pt}
%\centering
%\begin{tabular}{llllllllll} \hline
% & \multicolumn{2}{c}{$1^{st}$ cohort} & & \multicolumn{2}{c}{$2^{nd}$ cohort} &  & \multicolumn{2}{c}{total} & \\ \cline{2-3}\cline{5-6}\cline{8-9}
% & slide & bbox & & slide & bbox & & slide & bbox & \\ \hline
%train & 36 & 15606 & & 68 & 2375 & & 104 & 17981 & \\ \hline
%val & 13 & 1308 & & 21 & 685 & & 34 & 1993 & \\ \hline
%\end{tabular}
%\end{table}

\begin{table}[t]
\caption{Two Settings of train, validation and test in the classification task. Set 1 is divided by the scanner, and Set 2 is divided by if data is seen/unseen/mixed for model.}
\label{clsdata-split}
\centering
\label{detdata-split}
\begin{tabular}{lllllllll} \hline
\multirow{2}*{Group} & \multicolumn{3}{c}{$pos.$ slides} & & \multicolumn{3}{c}{$neg.$ slides} &\\ \cline{2-4}\cline{6-8}
& train & val & test & & train & val & test &\\ \hline
Set 1 & & & & & & & &\\\hline
$S_1$ & 88 & 7 & 8 & & 90& 8 & 7 &\\
$S_2$ & 242 & 18 & 20 & & 412 & 34 & 32 &\\
$S_3$ & 371 & 26 & 29 & & 270 & 20 & 18 &\\
$S_4$ & 109 & 8 & 7 & & 171 & 12 & 12 &\\\hline
Set 2 & & & & & & & &\\\hline
seen & 225 & 18 & 20 & & 414& 35& 33 &\\
unseen & 585 & 41 & 44 & & 529 & 39 & 36 &\\
mixed & 810 & 59 & 64 & & 943 & 74 & 69 &\\ \hline
%121test & 0 & 0 & 71 & 0 & 0 & 50\\\hline
\end{tabular}
\end{table}

The private 2,019 slides dataset consists of nine cohorts since there are different individuals produced in different circumstances. The local annotations at the patch-level are derived from WSIs among $1^{st}$ and $2^{nd}$ cohorts as shown in Table.\ref{datasets} with a limited amount. For the final WSI classification, all 2,019 slides are split to test the performance.

Slide data in our experiments are averagely split for train, validation and test of deep learning models with 85\%:7.5\%:7.5\%, as listed in Table.\ref{clsdata-split}. Because our final goal is WSI classification, there is no test set for CNN detectors/classifiers but only validation set (15\% of the whole data) to choose the best model.
%The recommended skeptical cells of the 138 slides are manually checked for the assessment of experiments (3).
Train, validation and test data of classification tasks are set in 2 ways as listed in Table.\ref{clsdata-split}.

In addition to the diagnosis label of each group is slightly uneven caused by the habit of individual experts, the cervical content of WSI varies with the unique parameter setting of 4 different scanners mentioned in Section 2. The division by different scanners is "Set 1" hence.

Besides, the discrepancy between nine cohorts is considered to perform the cervical feature representation of CNN detectors learned from the seen data, that is the local annotation among $1^{st}$ and $2^{nd}$ cohorts. Otherwise, the unseen data consists of other cohorts has never been learned for local recognition. It is more challenging for the generalization of CNNs. The division by seen/unseen/mixed is "Set 2" hence.

Each annotation is used once in each training epoch in every task. For training detectors, the bounding box is restricted inside the input image. It is the same for training classifiers, and the input image will be restricted inside the bounding box if the box size is larger than the image.

\subsubsection{Public data}

A public dataset by \cite{plissiti2018sipakmed} is used in our experiments to further show the significance of the proposed CNN. This dataset is divided into five classes and consists of total 4,049 cell images.

Following their experiments for patch-level classification task, we process a 5 cross-validation and run the best model (VGG19) for the comparison. Except for the input size is $80\times80$ as the same as the literature, other details of training like learning rate are the same as our CNN classifiers.

\section{Experimental Results}

There are two types of experimental results to illustrate the efficiency and the significance of the proposed method:
\begin{enumerate}[(1)]
\item The WSI classification for cervical diagnosis. %: Computing the score of cytopathological determination at the WSI-level.

\item The patch detection/classification for the lesion area. %: Computing the bounding box of the lesion zone in preselected patches.

%\item The recommendation for the lesion location with manually checking: Computing the top-$N$ skeptical lesion location of each WSI has patches learned.
\end{enumerate}

\subsection{Evaluation metrics}

The common estimation methods are used for the assessment. The true positive ($TP$), false positive ($FP$), true negative ($TN$), and false negative ($FN$) are the basic evaluation of results, thus the computations are as follows.
\begin{enumerate}[(1)]
\item Accuracy (Acc.), Sensitivity (Sens.), Specificity (Spec.), and the Receiver Operating Characteristic Curve (ROC) for classification tasks, as well as the nonparametric bootstrapping with 1,000 samples, was used to compute 95\% confidence intervals.

\begin{equation}\label{cls-metrics}
\begin{aligned}
  \text{Acc.} &= \frac{TP}{2(TP+FP)} + \frac{TN}{2(TN+FN)},\\
  \text{Sens.} &= \frac{TP}{TP+FN},\\
  \text{Spec.} &= \frac{TN}{TN+FP}.
\end{aligned}
\end{equation}

\item Precision (Prec.), Recall (Rec.), and mean of average precision in the precision-recall curve (mAP) for detection tasks.

\begin{equation}\label{det-metrics}
\begin{aligned}
  \text{Prec.} &= \frac{TP}{TP+FP},
  \text{Rec.} &= \frac{TP}{TP+FN}.
\end{aligned}
\end{equation}

%\item Hit-Rates for recommendation tasks by considering a circular area of radius $r=7.3 \mu m$ centered on each prediction as to the valid region for the detected lesion.

\end{enumerate}

\subsection{Implement details}

The whole experiment is implemented by PyTorch based on Windows 10 OS with one Xeon(R) 6134 CPU@3.20GHz and one Tesla P40 GPU. All models are training with the Adam optimizer and the Cosine learning rate reduction schedule. Aside from the code of ARF is obtained from github, other models are implemented by torchvision package.

\subsection{Quantitative results at WSI-level}

%Two groups of CNNs are evaluated to complete the comparison at WSI-level by the magnitude of the input side of feature extracting methods:
%\begin{enumerate}[Group 1]
% \item uses $2080\times2080\times3$, the input size of YOLOv3, which is the SOTA detection model with the relatively larger weight in the common method that has more repeated computations, 1,276 times with $O=288$.
% \item uses $6944\times6944\times3$, $8320\times8320\times3$ or $10400\times10400\times3$, where two smaller input sizes for MNV2 and InCNet$_{n_g=8}$, respectively, and the biggest for Tiny, YOCO and YOLCO. They are five lightweight CNNs with fewer repeated computations and $O=0$.
%\end{enumerate}

% Since the main idea of this study is based on the Tiny version of YOLOv3, these two CNNs as the contrast group are simultaneously tested for main results. MNV2 is the SOTA lightweight model, and InCNet$_{n_g=8}$ and YOCO are counterparts of the proposed YOLCO, as the contrast group to show the performance of the dual feature.

%\subsubsection{Main results}

\begin{table*}[h]\footnotesize
  \centering
  \caption{Results per scanning type of WSIs: Classification performance using Transformer classifier in Set 1, where AFR is the method in \cite{CAO2021102197}}\label{cls-scanner}
  %\begin{tabular}{lp{0.8cm}p{0.8cm}p{0.8cm}p{0.1cm}p{0.8cm}p{0.8cm}p{0.8cm}p{0.1cm}p{0.8cm}p{0.8cm}p{0.8cm}p{0.1cm}p{0.8cm}p{0.8cm}p{0.8cm}p{0.1cm}}
  \begin{tabular}{lp{0.6cm}p{0.0001cm}p{0.7cm}p{0.7cm}p{0.7cm}p{0.0001cm}p{0.7cm}p{0.7cm}p{0.7cm}p{0.0001cm}p{0.7cm}p{0.7cm}p{0.7cm}p{0.0001cm}p{0.7cm}p{0.7cm}p{0.7cm}p{0.0001cm}}
    \hline
     & AFR & & \multicolumn{3}{c}{Tiny}  & & \multicolumn{3}{c}{MNV2} & & \multicolumn{3}{c}{YOLOv3} && \multicolumn{3}{c}{YOLCO} &  \\\cline{2-2}\cline{4-6}\cline{8-10}\cline{12-14}\cline{16-18}
     & Acc. & & Acc.  & Sens.  & Spec. & & Acc.  & Sens. & Spec. & & Acc.  & Sens.  & Spec. & & Acc.  & Sens.  & Spec. & \\\hline
     All $S$ & .603 & & .65$\pm$.03 & .60$\pm$.13 & .70$\pm$.09 && {.70$\pm$.02} & .66$\pm$.13 & .74$\pm$.12 && .78$\pm$.02 & .84$\pm$.04 & .71$\pm$.06 && {.78$\pm$.03} & .86$\pm$.04 & .69$\pm$.08 & \\
     $S_1$ & .652 & & .74$\pm$.06 & .85$\pm$.13 & .62$\pm$.16 && {.82$\pm$.06} & .73$\pm$.09 & .90$\pm$.07 && .84$\pm$.03 & .83$\pm$.06 & .86$\pm$.00 && {.80$\pm$.04} & .80$\pm$.13 & .80$\pm$.07 & \\
     $S_2$ & .672 & & .62$\pm$.01 & .51$\pm$.12 & .73$\pm$.11 && {.70$\pm$.02} & .60$\pm$.10 & .80$\pm$.07 && .66$\pm$.02 & .66$\pm$.14 & .66$\pm$.10 && {.90$\pm$.03} & .97$\pm$.04 & .83$\pm$.04 & \\
     $S_3$ & .662 & & .64$\pm$.03 & .53$\pm$.22 & .75$\pm$.24 && {.71$\pm$.02} & .69$\pm$.11 & .73$\pm$.09 && .76$\pm$.01 & .74$\pm$.12 & .78$\pm$.13 && {.82$\pm$.04} & .79$\pm$.09 & .86$\pm$.06 & \\
     $S_4$ & .583 & & .79$\pm$.02 & .74$\pm$.15 & .85$\pm$.12 && {.60$\pm$.07} & .52$\pm$.30 & .68$\pm$.39 && .95$\pm$.03 & .94$\pm$.07 & .97$\pm$.04 && {1.0$\pm$.00} & 1.0$\pm$.00 & 1.0$\pm$.00 & \\\hline
     {AVG} & 0.634 &  & 0.688 &  &  & & 0.705 &  &  & & {0.798} &  &  & & \textbf{0.862} \\
     {STD} & 0.035 &  & {0.074} &  &  & & {0.080} &  &  & & {0.098} &  &  & & {0.089} \\
     {MD} & 0.089 &  & {0.236} &  &  & & {0.438} &  &  & & {0.366} &  &  & & {0.271} \\
    \hline
  \end{tabular}
\end{table*}

\begin{table*}\footnotesize
  \centering
  \caption{Results on seen, unseen, and mixed WSIs: Classification performance using Transformer classifier in Set 2, where AFR is the method in \cite{CAO2021102197}}\label{cls-familiarity} % YOLCO.x denotes YOLCO with the extra feature vectors collected from the potential hard negative locations while its $N$ increased up to double to others.
  %\begin{tabular}{lp{0.8cm}p{0.8cm}p{0.8cm}p{0.1cm}p{0.8cm}p{0.8cm}p{0.8cm}p{0.1cm}p{0.8cm}p{0.8cm}p{0.8cm}p{0.1cm}p{0.8cm}p{0.8cm}p{0.8cm}p{0.1cm}}
  \begin{tabular}{lp{0.6cm}p{0.0001cm}p{0.7cm}p{0.7cm}p{0.7cm}p{0.0001cm}p{0.7cm}p{0.7cm}p{0.7cm}p{0.0001cm}p{0.7cm}p{0.7cm}p{0.7cm}p{0.0001cm}p{0.7cm}p{0.7cm}p{0.7cm}p{0.0001cm}}
    \hline
     & AFR & & \multicolumn{3}{c}{Tiny} &  & \multicolumn{3}{c}{MNV2} &  & \multicolumn{3}{c}{YOLOv3} & & \multicolumn{3}{c}{YOLCO} & %\\\cline{2-4}\cline{6-8}\cline{10-12}\cline{14-16}
     \\\cline{2-2}\cline{4-6}\cline{8-10}\cline{12-14}\cline{16-18}
     & Acc. & & Acc. & Sens. & Spec. & & Acc. & Sens. & Spec. & & Acc. & Sens. & Spec. & & Acc. & Sens. & Spec. &\\\hline% & Acc. & Sens. & Spec. \\\hline
     seen & .723 & & .62$\pm$.01 & .42$\pm$.07 & .82$\pm$.08 && {.82$\pm$.01} & .86$\pm$.03 & .77$\pm$.03 && .80$\pm$.02 & .79$\pm$.07 & .81$\pm$.06 && {.89$\pm$.01} & .87$\pm$.05 & .90$\pm$.03 & \\
     unseen & .542 & & .61$\pm$.02 & .78$\pm$.10 & .45$\pm$.13 && {.66$\pm$.01} & .47$\pm$.14 & .86$\pm$.13 && .78$\pm$.01 & .71$\pm$.10 & .84$\pm$.09 && .72$\pm$.04 & .70$\pm$.22 & .73$\pm$.16 & \\
     mixed & .603 & & .66$\pm$.02 & .65$\pm$.09 & .68$\pm$.08 && {.70$\pm$.02} & .66$\pm$.13 & .74$\pm$.12 && .78$\pm$.02 & .84$\pm$.04 & .71$\pm$.06 && {.78$\pm$.03} & .86$\pm$.04 & .69$\pm$.08 & \\\hline
     {AVG} & 0.622 & & 0.630 & & & & 0.726 & & & & {0.786} & & & & \textbf{0.797} \\
     {STD} & 0.075 & & {0.027} & & & & {0.067} & & & & \textbf{0.021} & & & & {0.075} \\
     {MD} & 0.181 & & {0.108} & & & & {0.188} & & & & \textbf{0.084} & & & & {0.242} \\
    \hline
  \end{tabular}
\end{table*}

\begin{table}
  \centering
  \caption{Performance of YOLCO (mixed) on positive subclasses of $8^{th}$ cohort WSIs.}
  \label{exp:subclass}
  %\begin{tabular}{lllllllllllll}
  \begin{tabular}{p{1.75cm}p{1.25cm}p{1.25cm}p{1.25cm}p{1.25cm}}
    \hline
    % & ASC-US & ASC-US & ASC-US &  & LSIL & LSIL & LSIL & & HSIL & HSIL & HSIL & \\\cline{2-4}\cline{6-8}\cline{10-12}
    % & train & val & test & & train & val & test & & train & val & test &  \\\hline
    %\multicolumn{4}{l}{Private data ($n=235$)}\\\hline
     & $neg.$ & ASC-US & LSIL & HSIL \\
     & $n=144$ & $n=39$ &$n=27$ &$n=25$ \\\hline
    Accuracy & 0.837 & 0.849 & 0.753 & 0.786  \\
    Sensitivity & 0.933 & 0.867 & 1.000 & 1.000 \\
    Specificity & 0.740 & 0.831 & 0.507 & 0.571  \\\hline
%    \multicolumn{4}{l}{Public data (patch, $n=4,049$)} \\\hline
%    Accuracy & 0. & 0. & 0.  \\
%    Sensitivity & 0. &  &  \\
%    Specificity & 0. & 0. & 0.  \\\hline
%    \multicolumn{4}{l}{Public data (patch, $n=963$)} \\\hline
%    Accuracy & - & 0. & 0.  \\
%    Sensitivity & - &  &  \\
%    Specificity & - & 0. & 0.  \\\hline
  \end{tabular}
\end{table}

\begin{table*}[ht]
  \centering
  \caption{Results per classifier: Classification performance on mixed scanners data (2,019 slides), where "Trans." denotes Transformer model.}\label{cls-classifier}
  \begin{tabular}{lllllllllllllllll}
    \hline
     \multirow{2}*{} & \multicolumn{3}{c}{Tiny}  & & \multicolumn{3}{c}{MNV2} & & \multicolumn{3}{c}{YOLOv3} & & \multicolumn{3}{c}{YOLCO} &  \\\cline{2-4}\cline{6-8}\cline{10-12}\cline{14-16}
     & Acc.  & Sens.  & Spec. & & Acc.  & Sens. & Spec. & & Acc.  & Sens.  & Spec. & & Acc.  & Sens.  & Spec. & \\\hline
     SVM & 0.638 & 0.609 & 0.667 & & 0.516 & 0.047 & 0.986 & & 0.585 & 0.547 & 0.623 & & 0.726 & 0.641 & 0.811 & \\
     RNN & \textbf{0.659} & 0.781  & 0.536 & & {0.500} & 1.000 & 0.000 & & 0.526 & 0.922 & 0.130 & & 0.787 & 0.734 & 0.841 & \\
     LSTM & 0.653 & 0.828 & 0.478 & & {0.500} & 1.000 & 0.000 & & 0.787 & 0.922 & 0.652 & & 0.745 & 0.766 & 0.725 & \\
     Trans. & {0.601} & 0.391 & 0.812 & & \textbf{0.725} & 0.813 & 0.638 & & \textbf{0.806} & {0.859} & {0.754} & & \textbf{0.808} & 0.906 & 0.710 & \\
    \hline
  \end{tabular}
\end{table*}

%    SVM
%     ['mnv2disconserve0-top100']
%acc, [0.5263157894736843]
%tpr, [0.5488721804511278]
%tnr, [0.9774436090225564]

% \subsubsection{Performance on Set 1}

Experiences in Set 1 show the performance of methods for different scanning devices. As shown in Table.\ref{cls-scanner}, the proposed YOLCO has the highest average accuracy (AVG), $86.2\%$, and the lower fluctuation of all data groups. It precedes the second-best YOLOv3 with $6.4\%$ higher accuracy as well as the robustness with $0.009$ lower STD (standard deviance) and $0.095$ lower MD (max difference). Despite the subtle lower STD of Tiny and MNV2, and lower MD of Tiny than YOLCO, the much lower accuracy is unacceptable to diagnose cervical slides, which is mainly affected by that it has just surpassed $60\%$  accuracy in scanner $S_2$ of Tiny or in scanner $S_4$ of MNV2. A similar hardship occurs to YOLOv3 in that the performance in $S_2$ is unmatched with remains. YOLCO is considerably stable and efficient than common models, with the maximum accuracy in $S_4$ and more powerful in the above hardship, based on the feature representation enhanced by InCNet.

% \subsubsection{Performance on Set 2}

As shown in Table.\ref{cls-familiarity}, we get the generalization of each method when it is test on unseen or mixed data in Set 2. YOLCO with $79.7\%$ average accuracy has the highest precision than others but the most susceptible whether on seen/unseen/mixed data with the highest MD $0.242$. It is reasonable that the deep learning network has the best capability for seen cervical slides, has the worst for unseen cervical slides, and the mixed data is in the middle. The existent pattern of this statement is suspicious of YOLOv3 with the similar accuracy of unseen and mixed data, and it shows the norm of the ability of YOLCO that it is limited when feature extraction is learned from relatively few amount of annotations and few amount of parameters.

\begin{figure*}[!t]
  \centering
  \includegraphics[width=1.01\linewidth]{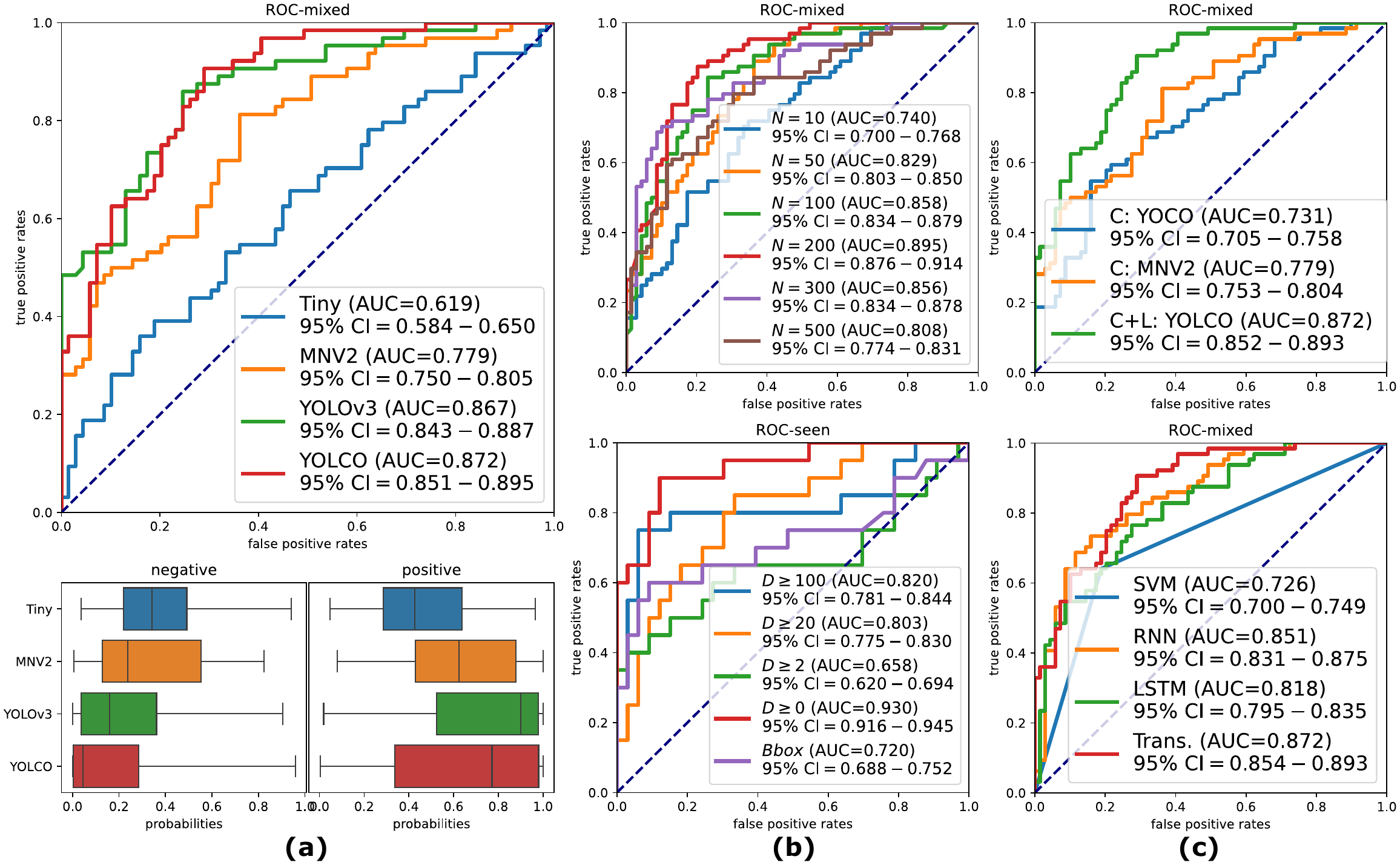}
  \caption{The quantitative results of ROC curves and probabilities distribution, where the last run of loops for the sequence classification model is illustrated for each method, and 95\% CI is the 95\% confidence interval. (a) shows the overall performance on our 2,019 WSIs data using Transformer classifier; (b) shows the effect of parameters changing for the proposed method by controlled variables, where "Bbox" denote collecting method ii, $N$ varies with $D=0$, and $D$ varies with $N=100$; and (c) shows the WSI-level comparison between multiple tasks and the single task at the top, and the usage of different classifiers of our method at the bottom.}\label{roc-box}
\end{figure*}

% Overall, the comparisons of accuracy for the WSI-level classification as shown in Table.\ref{cls-scanner} and Table.\ref{cls-familiarity} in all of our data can validate the superiority of our method over the best of common CNN.
As for the further revealing, the ROC curve as shown in the top of Figure.\ref{roc-box} (a) from the last run of loops in our mixed data denotes not only classification accuracy but the probabilities provided by YOLCO are more accurate than the performance of common CNNs. The higher area under the curve (AUC) score also quantitatively shows our method more accurate than the best of others: 0.872 of YOLCO vs 0.867 of YOLOv3. And the box chart at the bottom displays the distribution of probabilities that our method performs better.

\subsubsection{Performance on positive subclasses}

Since the clinical concern is more lie on the high sensitivity of severe types, like HISL, we show such results on our $8^{th}$ cohort slides in Table \ref{exp:subclass}. %, and a public multi-class datasets \cite{hussain2020liquid}

%The first public dataset by \cite{plissiti2018sipakmed} consists of 4,049 cropped cells from 966 ROIs. The input to our InCNet$_{n_g=8}$ is cropped cell images of this dataset, where the data size is various. Thus, we resized them to a fixed size (follow the literature, 80 × 80 pixels each).
%
%Another one consists of 963 ROIs with the same size of $2048\times1536$ pixels.

% The public dataset consists of 963 ROIs with the same size of $2048\times1536$ pixels. There are 4 classes: NILM, LSIL, HSIL, and SCC. The proposed model InCNet$_{n_g=8}$ is trained follow the literature.

\subsubsection{Ablation studies: sequence classifiers}
\label{sec:exp-classifier}

Comparing with YOLOv3, Tiny, MNV2, and our method, different classifiers are adopted into our framework for 2,019 WSIs as shown in Table.\ref{cls-classifier}. And, Figure.\ref{roc-box} (c) bottom is ROC results of our method in the last run of repeated experiments.

Our method, YOLCO, displays the best performance on every classifier except LSTM from $72.6\%$ to $80.8\%$. In contrast, a striking gap between accuracies of the second, YOLOv3 from $58.5\%$ to $80.6\%$, indicates its inferior feature representation. Furthermore, even though YOLOv3 has $78.7\%$ accuracy based on a relatively more complex classifier, the proposed method is superior using the non-parametric SVM.

%In consideration of the principles of SVM and other sequence classifiers, the performance of SVM directly reveals the inherent ability of each CNN encoder, and there is higher accuracy when more parameters of the sequence classifier, aside from Tiny and MNV2 the weakest model and the failed feature representation, respectively.

\subsubsection{Ablation studies: collecting methods}

As shown in Figure.\ref{roc-box} (b), it is apparent that more features from cervical slides lead to better performance of feature representation, as well as the AUC score improved from $0.740$ to $0.895$ when $N$ is increased from $10$ to $200$. It is fallen when more boxes are chosen, however, since the long sequence classification is a challenging task.

A glance at the bottom ROC figure reveals the features are valuable and rich between the neighboring patch that $0.93$ the best AUC is achieved via $D=0$. An another evidence is that $D=2$ leads the worst results, where the edge of abnormal cells have more chance to be collected than the center cause of $size$ $of$ $nucleus$ $<D<$ $size$ $of$ $cell$. The weak but not the worst method (ii) \textsl{Bbox} hardly reaches a better performance.

%In conclusion, $D=0, N=100$ are the optimal parameter setting in our method.

\begin{table*}\footnotesize
  \centering
  \caption{Results of ablation study: Classification performance at the patch-level and the WSI-level. InCNet is set with ${n_g=8}$ here, and $P.$ means the weight of parameters of network. }\label{is-ablation} % , where "C" denotes that CNN is supervised by $\mathcal L_{cls}$, and "L" denotes that CNN is supervised by $\mathcal L_{box}$
  \begin{tabular}{p{0.4cm}p{0.95cm}p{0.95cm}p{0.95cm}p{0.95cm}p{0.0001cm}p{0.95cm}p{0.95cm}p{0.95cm}p{0.95cm}p{0.0001cm}p{0.95cm}p{0.95cm}p{0.95cm}p{0.95cm}}
    \hline
     & \multicolumn{4}{c}{patch-level ($1^{st}$ \& $2^{nd}$ cohorts)} & & \multicolumn{4}{c}{WSI-level ($1^{st}$ \& $2^{nd}$ cohorts)} & & \multicolumn{4}{c}{WSI-level (ALL cohorts)} \\\cline{2-5}\cline{7-10}\cline{12-15}
     & InCNet & MNV2 & YOCO & YOLCO & & InCNet & MNV2 & YOCO & YOLCO & & InCNet & MNV2 & YOCO & YOLCO  \\\hline
     Loss & $\mathcal L_{cls}$ & $\mathcal L_{cls}$ & $\mathcal L_{cls}$ & $\mathcal L_{cls}\&\mathcal L_{box}$ & & $\mathcal L_{cls}$ & $\mathcal L_{cls}$ & $\mathcal L_{cls}$ & $\mathcal L_{cls}\&\mathcal L_{box}$ & & $\mathcal L_{cls}$ & $\mathcal L_{cls}$ & $\mathcal L_{cls}$ & $\mathcal L_{cls}\&\mathcal L_{box}$  \\
     $P.$ & \cellcolor[gray]{0.8}$2.79M$ & \cellcolor[gray]{0.7}$2.23M$  & \cellcolor[gray]{0.6}$1.79M$  & $1.79M$ & & \cellcolor[gray]{0.8}$2.79M$ & \cellcolor[gray]{0.7}$2.23M$  & \cellcolor[gray]{0.6}$1.79M$  & $1.79M$ & & \cellcolor[gray]{0.8}$2.79M$ & \cellcolor[gray]{0.7}$2.23M$  & \cellcolor[gray]{0.6}$1.79M$  & $1.79M$\\\hline
%    Sens. & .96$\pm$.01 & .96$\pm$.01 & .82$\pm$.01 & .74$\pm$.01 & & .83$\pm$.04 & .86$\pm$.04 & .63$\pm$.03 & .86$\pm$.05 & & .66$\pm$.04 & .66$\pm$.13 & .71$\pm$.11 & .86$\pm$.04 & \\
%    Spec. & .97$\pm$.00 & .96$\pm$.01 & .96$\pm$.01 & .97$\pm$.00 & & .89$\pm$.02 & .76$\pm$.03 & .96$\pm$.04 & .91$\pm$.03 & & .67$\pm$.06 & .74$\pm$.12 & .67$\pm$.12 & .69$\pm$.08 & \\
%    Acc. & .96$\pm$.00 & {.96$\pm$.00} & .89$\pm$.01 & .86$\pm$.01 & & .86$\pm$.02 & .82$\pm$.01 & .79$\pm$.02 & {.89$\pm$.01} & & .67$\pm$.01 & .70$\pm$.02 & .69$\pm$.01 & {.78$\pm$.03} & \\
%    AUC & \textbf{.964$\pm$.004} & {.959$\pm$.004} & .889$\pm$.005 & .855$\pm$.006 & & .890$\pm$.007 & .862$\pm$.009 & .801$\pm$.019 & \textbf{.927$\pm$.011} & & .692$\pm$.004 & .746$\pm$.018 & .750$\pm$.015 & \textbf{.841$\pm$.030} & \\
    Sens. & .957$\pm$.007 & .958$\pm$.006 & .820$\pm$.009 & .740$\pm$.014 & & .825$\pm$.038 & .860$\pm$.037 & .625$\pm$.025 & .860$\pm$.049 & & .664$\pm$.043 & .659$\pm$.129 & .709$\pm$.113 & .859$\pm$.040  \\
    Spec. & .971$\pm$.004 & .961$\pm$.005 & .958$\pm$.008 & .970$\pm$.004 & & .894$\pm$.015 & .764$\pm$.030 & .962$\pm$.039 & .909$\pm$.027 & & .669$\pm$.058 & .739$\pm$.119 & .672$\pm$.121 & .693$\pm$.080  \\
    Acc. & .964$\pm$.004 & {.959$\pm$.004} & .889$\pm$.005 & .855$\pm$.006 & & .860$\pm$.015 & .822$\pm$.009 & .794$\pm$.021 & {.886$\pm$.013} & & .667$\pm$.011 & .699$\pm$.017 & .691$\pm$.008 & {.776$\pm$.030}  \\
    \textbf{AUC} & \cellcolor[gray]{0.8}{.964$\pm$.004} & \cellcolor[gray]{0.7}{.959$\pm$.004} & \cellcolor[gray]{0.6}.889$\pm$.005 & .855$\pm$.006 & & \cellcolor[gray]{0.8}.890$\pm$.007 & \cellcolor[gray]{0.7}.862$\pm$.009 & \cellcolor[gray]{0.6}.801$\pm$.019 & {.927$\pm$.011} & & \cellcolor[gray]{0.6}.692$\pm$.004 & \cellcolor[gray]{0.7}.746$\pm$.018 & \cellcolor[gray]{0.8}.750$\pm$.015 & {.841$\pm$.030}  \\
    \hline
  \end{tabular}
\end{table*}

\subsubsection{Dual tasks: classification + location}

To illustrate the significance of the proposed feature mixture, the WSI-level ablation study is executed by comparing between performances of CNNs producing single feature under the single supervision, the classification, or feature mixture under the dual task, classification + location.

% MNV2 and InCNet$_{n_g=8}$ are the SOTA lightweight model and the counterpart with the same architecture, respectively, which are trained under the classification task at the patch-level. And, YOCO is the strict contrast model that has the same architecture and train-val data of YOLCO, but only deducts the box loss ($\mathcal L_{box}$).

In the Table.\ref{is-ablation}, and the top of Figure.\ref{roc-box} (c), given are performances of classification at patch-level with the equivalent number of negative patches, and performances at WSI-level, respectively. %Note that the patch-level data are randomly sampled inside the box annotation during the evaluation, the experience is repeated ten times hence.
Reasonably, the AUC score has $0.034$ up at the path level classification after the box loss deducting in YOCO but has $0.091$ down that feature mixture is more valid at the WSI-level based on the proposed idea. And, InCNet$_{n_g=8}$ achieves $0.964$ a remarkable AUC at patch-level with $0.005$ higher than the SOTA model and others, as well as the same cohorts at WSI-level with $0.89$ AUC. But, it becomes $0.722$ the lowest at the WSI-level as similar as remains without $\mathcal L_{box}$ for all cohorts including unseen data.

\subsection{Feature Representation Qualitative results}

\subsubsection{Feature variety}

\begin{figure}[!t]
  \centering
  \includegraphics[width=\linewidth]{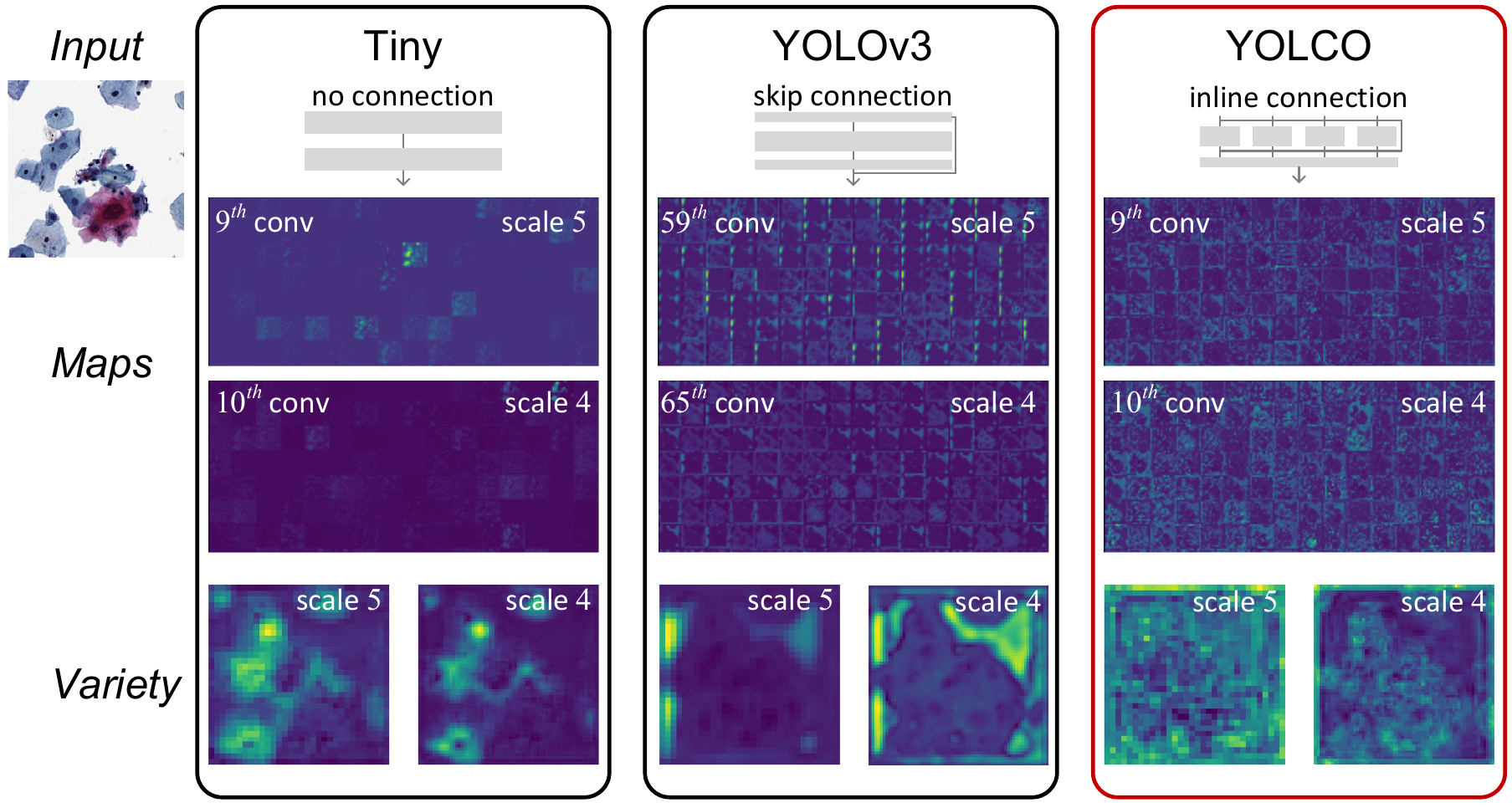}
  \caption{The feature variety of the proposed model against two other conventional CNNs. All feature maps are extracted from the last layer before the output layer at each level of scale with the presented input image, where the downsampling ratio is $2^x$ with scale $x$. The variety is the channel-wise standard deviation of feature maps.}\label{cnn_interpret}
\end{figure}

The feature variety is depicted via the visualization of the feature map of CNNs as shown in Figure.\ref{cnn_interpret}, to illustrate the weakness of natural upstream CNNs and the advantage of the proposed YOLCO. The activated part in every feature map in the figure has a higher value than the non-activated value, which is less than or equals 0 based on the LeakyRELU function. And, both Tiny, YOLOv3, and YOLCO generate feature maps at multiple scales from the $1024\times1024$ RGB input image.

In the maps of the figure, with no connection, Tiny using the same build of convolutional modules as YOLCO has the fewest activated feature maps, which consist of more than half of nearly non-activated. With skip connection, YOLOv3 using the much deeper structure and bigger parameters can generate much more feature maps (1792 channels). It performs an overwhelming similar pattern of activation but much better than its downgraded version, Tiny.

In the variety of the figure, the natural upstream CNNs, Tiny and YOLOv3, are both have a similar pattern of variety of their feature maps at two scales, while YOLCO has the most various feature maps. High values of upstream CNNs are existed but not in the most important abnormal cells on the lower and central parts of the input image. That is the weakness for subsequent WSI classification based on those features.

\subsubsection{t-SNE}

To illustrate the discrimination and the feature representation between CNNs, the WSI-level features are embedded into the 2D plane using the t-SNE toolbox (\cite{van2008visualizing}) with the same random seed (142857). SVM, LSTM, and Transformer results are point in the t-SNE results from the $ n_c*N$ dimensional features for SVM from CNNs, and the features for LSTM and Transformer from the last hidden state, respectively.

\begin{figure*}[!t]
  \centering
  \includegraphics[width=\linewidth]{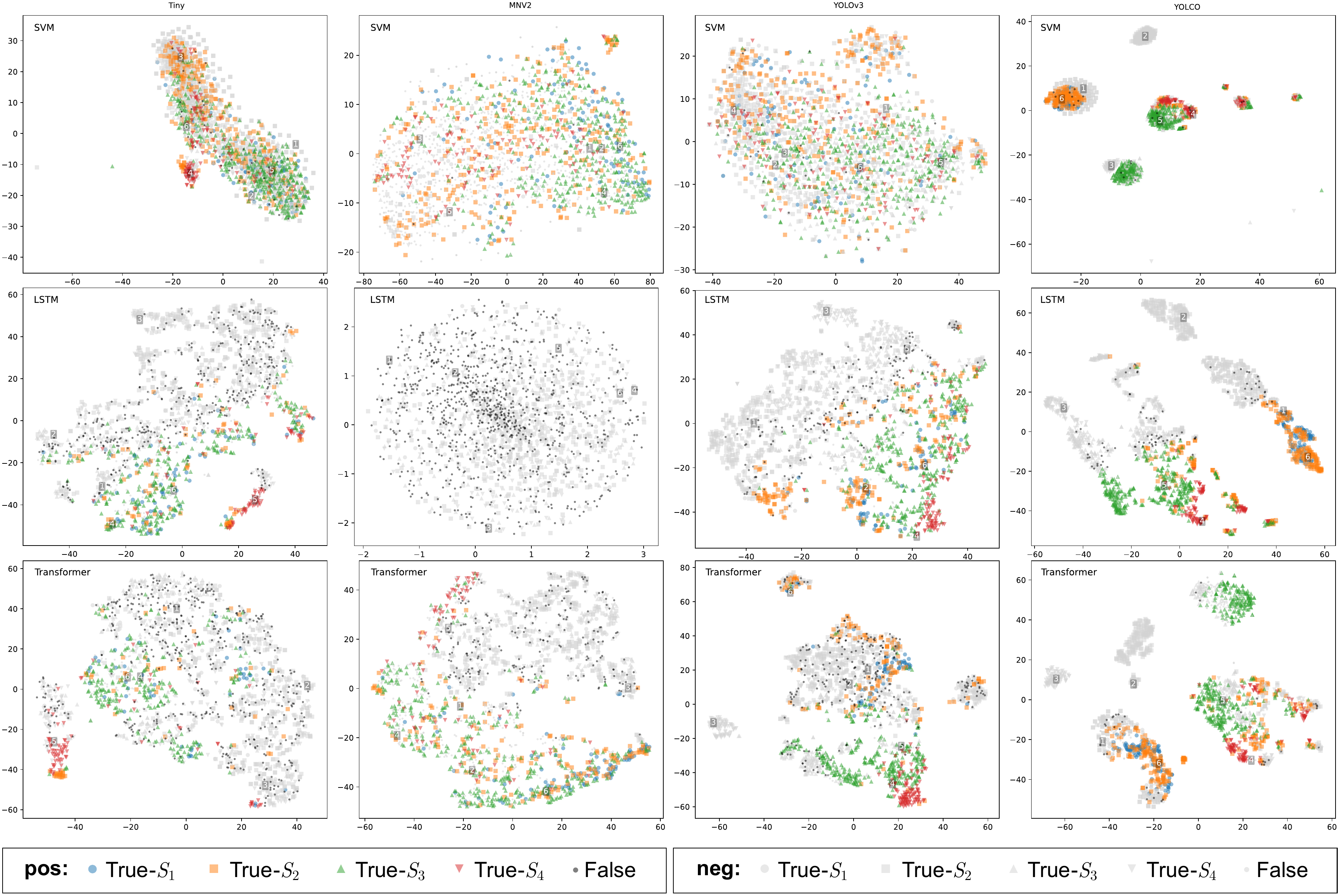}
  \caption{Feature embedding from raw feature vectors (SVM), and embedded feature vectors (LSTM and Transformer) using t-SNE toolbox (\cite{van2008visualizing}).}\label{tsne-ana}
\end{figure*}

As shown in Figure.\ref{tsne-ana}, it is clear to found that the $S_2$ data are the most challenging scanner type that is the most scattered. In comparison to the CNN feature representation, t-SNE results in SVM of the proposed YOLCO are clearly distributing in 7 clusters, against the ambiguous results by common CNNs. Positive slides and negative slides are gradually clustered more closer with classifiers from LSTM to Transformer in our method. In contrast, slides are hard to distinct even the classifier becoming more complex aside from $S_4$ type. Additionally, 3 positive and 3 negative slides are randomly selected based on the distribution of results of the proposed method in SVM to observe the separability, and every slide is moved into its true group in YOLCO via the most powerful Transformer.

\subsubsection{Examples}

\begin{figure*}[!t]
  \centering
  \includegraphics[width=\linewidth]{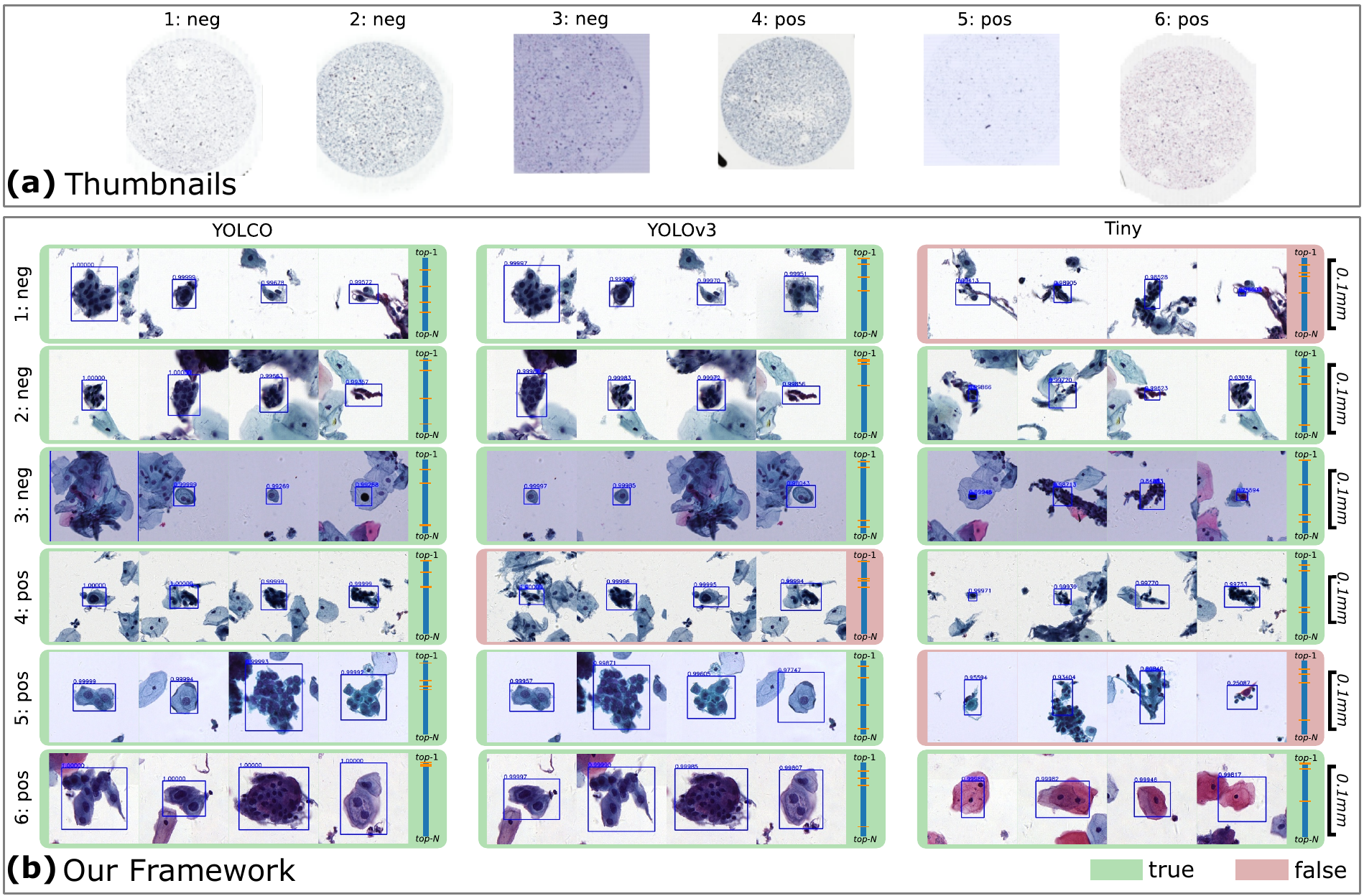}
  \caption{6 examples to illustrate the performance, where slides are chosen in t-SNE analysis (Figure.\ref{tsne-ana}), and "pos" denotes the positive diagnosis from the golden standard as well as "neg" denotes the negative diagnosis. (a) thumbnails at scale level 8 of 6 slides. (b) the intersection of 4 representative detection results between models under our framework, where Tiny usually performs different with others.}\label{10-slides} %(c) the representative 4 patch images with detection results of AttFPN-Faster-RCNN (AFR) under the traditional framework.
\end{figure*}

The representative examples of two different frameworks are shown in Figure.\ref{10-slides}. There are three detectors used in our framework and their representative detection of above 6 slides are shown, where No. 6 belongs to $S_1$, No. 1 and 2 belong to $S_2$, No. 4 belong to $S_3$, and No. 3 and 5 belong to $S_4$.

In Figure.\ref{10-slides} (b), we choose 4 of the intersection of top-100 detection results with the indication of their ranks. It is clear that YOLCO and YOLOv3 have a sufficient intersection, because they have similar accuracy in WSI classification and YOLCO has a slight superiority. However, ranks of the intersection are different. For instance, the highest detection result of YOLCO is ranked in the third place in No. 4 example, and negative cells are sorted ahead the positive cell. It is the reason why YOLOv3 gave a false prediction of No. 4 WSI. While Tiny usually output negative cells or artifacts, which are rarely existed in other better methods. That can explain why Tiny has a much lower accuracy than others.

%In the traditional framework \cite{CAO2021102197} in Figure.\ref{10-slides} (c), 4 representative patch images output by ResNet50 are shown with the detection results of AFR.

%Firstly, CNN is prone to recommend different lesion locations with others than the same as shown in Figure.\ref{10-slides} (a). Then, bounding box results of the proposed model have similar morphology with results of YOLOv3 in Figure.\ref{10-slides} (b), while some possibly positive cell clusters are avoided from YOLOv3 but YOLCO. Tiny has many uninterpretable detection results, however, that focus on the nucleus with high probabilities. Lastly, the fluctuation during the sequence classification in Figure.\ref{10-slides} (c) can further indicate CNN features from our method are more representative to determine WSIs that probabilities are almost still in No. 0, 2, 3, 7, and 8, are just slight fluctuated in No. 5 and 9, and strong fluctuated in No. 1, 4, and 6.

\subsection{Quantitative results at patch-level}
\label{sec:det-results}

\begin{table}
\caption{Results on patch-level data: Detection performance on the private dataset $n=19,974$. Classification performance on a public datasets by \cite{plissiti2018sipakmed}, $n=4,049$, where VGG19 is the best model of the literature.} %and COCO 2014, where FPS denotes the seconds cost per image patch ($1024\times1024$) in the train phase
\label{det-map}
\centering
\begin{tabular}{lllll} \hline
% & \multicolumn{3}{c}{Cervical} &  & \multicolumn{2}{c}{COCO 2014} & & \multicolumn{2}{c}{Costs} & \\\cline{2-4}\cline{6-7}\cline{9-10}
% & mAP$_{.5}$ & Prec. & Rec. & & mAP$_{.5}$& mAP$_{.5:.95}$ & & Param. & FPS$_{1024}$ & \\ \hline
 %& \multicolumn{3}{c}{Private} &  \\\cline{2-4}%& \multicolumn{2}{c}{Public} & \multirow{2}*{FPS$_{1024}$} \\\cline{2-4}\cline{6-7}
Private & Param. & mAP$_{.5}$ & Rec. & Prec. \\ \hline%& Acc. (1) & Acc. (2) & \\ \hline
%YOLOv5l & $0.379$ & $0.327$ & $0.589$ & & $\textbf{0.665}$ &  $\textbf{0.477}$ & {6.40 p/s} \\
%YOLOv4 & $\mathbf{0.789}$ & $0.271$ & $0.955$ & & \textbf{0.657} & \textbf{0.435} & & $\mathbf{20.6M}$ & 3.60 p/s& \\
YOLOv3 & 61.5M & $\underline{0.735}$ & $0.943$ & $0.352$\\%& ${0.579}$ & ${0.330}$ & {3.99} p/s \\ \hline
%YOLOv5s & $0.486$ & $0.603$ & $0.390$ & & $\textbf{0.554}$ & $\textbf{0.367}$ & {22.35 p/s} \\
Tiny & 8.67M & $0.674$ & $0.919$ & $0.261$\\%& $\textbf{0.348}$ & $\textbf{0.176}$ & 11.55 p/s \\
{YOLCO} & 1.79M & ${0.725}$ & \textbf{0.955} & $0.215$\\ \hline%& ${0.277}$ & ${0.141}$ & {12.48} p/s \\ \hline
 Public & & Acc. & Sens. & Spec. \\ \hline%& Acc. (1) & Acc. (2) & \\ \hline
 VGG19 & 144M & .9950 & .9978 & .9938 \\
 MNV2 & 2.23M & .9942 & .9981 & .9908\\
 InCNet$_{n_g=8}$ & 2.79M & \underline{.9952} & \textbf{.9984} & .9922\\\hline
 %Public 2 & & \\ \hline%& Acc. (1) & Acc. (2) & \\ \hline
\end{tabular}
\end{table}

%The assessment of the cytopathological data and the public data is accomplished to illustrate the lesion detection function.

\subsubsection{Private data}

For the private data, detection results of YOLO series and YOLCO are listed in Table.\ref{det-map}. It is evident that the proposed CNN is performing a great mAP $72.5\%$ and the best recall (Rec.) comparing with others. It shows a nice comparison with SOTA YOLOv3 with only $0.04\times\sim0.08\times$ weighting in parameters and $1.0\%\sim6.4\%$ decreasing in mAP.

\subsubsection{Public data}

For the public data, classification results of the best reported model VGG19, the SOTA model MNV2, and InCNet are listed in Table.\ref{det-map}. The significance of our model can be proved by not only the best average accuracy (Acc.) $99.52\%$ of all classes but also the best average sensitivity (Sens.) $99.84\%$.

\subsubsection{Ablation studies: connecting methods}
\label{sec:ablation-patch-level}
\begin{figure}[!t]
\centering
\includegraphics[width=.8\linewidth]{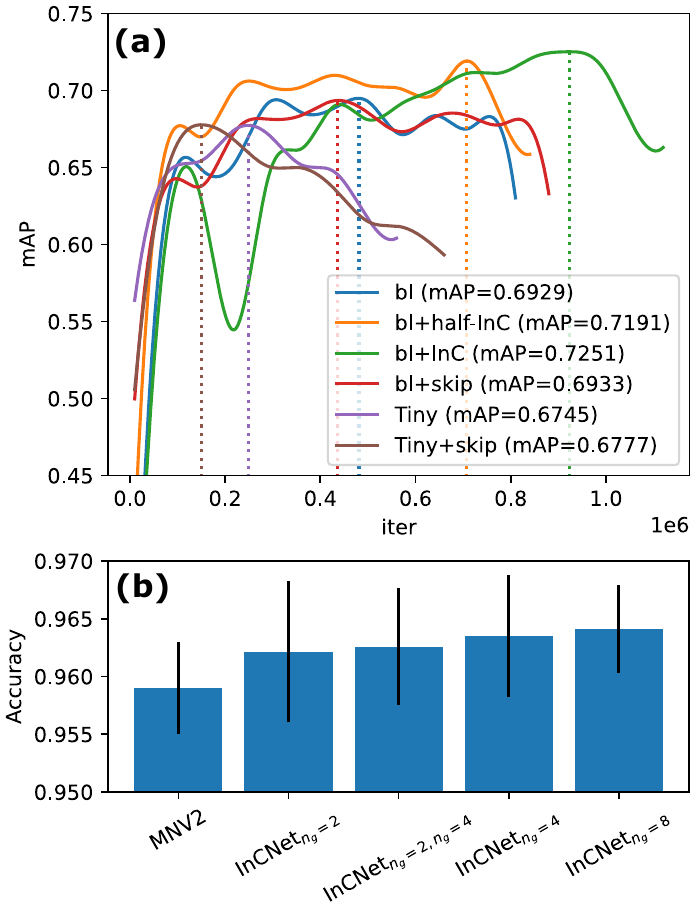}
\caption{Ablation studies for our inline connection at the patch-level. (a) The detection performance of proposed YOLCO (bl+InC) and its counterpart Tiny, where "half" denotes InC connected with only half of groups. (b) The classification performance of MNV2 and its varieties with the replacement of InverseResidual module by InCNet with the same channels, where the vertical line over the bar is the error bar from ten repeated runs, $n_g$ is the group number, and the first four modules consist of fewer groups when $n_g$ has two values.}
\label{patch-level-ablation}
\end{figure}

Ablation studies are accomplished to test the workability of connecting methods via changing the connectivity of Tiny, YOLCO, MNV2, and InCNet, which are based on the common convolution or the depthwise separable (DS) convolution. Note that "bl", in Figure \ref{patch-level-ablation} (a), is the baseline YOLCO that be without any connection, the "skip" connection is the skip/residual connection, and "half-InC" is the ablated version of the inline connection that only the first half groups are connected. And, in (b), different settings of $n_g$, the number of groups, show how the connectivity influences the classification performance.

Given carves in (a) that collected mAPs of validation data during the training phase of 6 CNNs reveal the model has a higher mAP with more connectivity. Especially, $67.4\%$ and $69.3\%$, the lower mAPs of Tiny and bl which are non-connected, respectively, rising to $67.8\%$ and $69.4\%$, the higher mAPs of Tiny+skip and bl+skip. In the meantime, except for the comparison between common convolutions from Tinys, bls are overfitted later and better with more connectivity, where $\text{skip}<\text{half-InC}<\text{InC} = 72.5\%$, the best performance from the proposed method bl+InC. Additionally, the patch-level accuracy rising with more connectivity in (b) can further illustrate the performance of InCNet for cytopathology.

\subsection{C++ Deployment}

Considering that the most efficient way to get computer-aided prediagnosis is to have a similar time cost between WSI computation and scanning. In python, our method is the fastest that needs 538.0s to compute one WSI, while the second best YOLOv3 needs 1352.6s (the proposal is $2.51\times$ faster). Although implementing the proposed method by python is friendly for showing and comparing the performance, the least time cost that is 538s per WSI is impractical. Thus, we release a C++ deployment of the proposed method in github to test the practical speed for application.

\begin{table} [!t]
\caption{The practical speed of the proposed method after C++ deployment for two common WSI formats, SVS and MRXS. The release executable file is in github, where GMem denotes the GPU memory needing.}
\label{deployment}
\centering
\begin{tabular}{lllll} \hline
&\multirow{2}*{GMem} & \multicolumn{2}{c}{Speed$_{s/\text{WSI}}$} \\\cline{3-4}
&&SVS&MRXS\\\hline
Libtorch & 8.8 GB & 69.9 & 70.1 & \\
Libtorch + half & 5.7 GB & 53.2 & 53.5 & \\\hline
\end{tabular}
\end{table}

As listed in the Table.\ref{deployment}, the proposed method is speeded up to $\sim$70 s per WSI whether it is SVS format or MRXS format after C++ deployment. It can be faster to $\sim$53 s per WSI using half precision (single float precision).

The bottleneck of speeds in python is the limited multiple thread programming and the dynamic computation graph of Pytorch. In C++, we translate the python codes into an efficient multiple thread program, and generate the static computation graph from Pytorch model by using Libtorch, which is the official C++ deployment solution for Pytorch. Using 4 data loaders, the time cost of data loading is reduced to average $\sim$100ms per tile image. And, the model inference needs only $\sim$1s per tile image with the static computation graph.

\section{Discussion}

The effective computation has been mainly explored on preprocessing or postprocessing instead of the rethought of the backbone in the past. Applying the extraction for the foreground or the ROI before the CNN encoding, however, generally reduces the half pixels at most in our data. A highly intergrated feature embedding is more efficient, and especially effective in the medical image field. According to this, recognizing ROI with the low magnification then sliding a small window of ROIs with the high magnification is the approach to represent the WSI by few small patch images. Some computations that are repeated for the same pixel with multiple models, however, are inevitable.

Thus, YOLCO (You Only Look Cytopathology Once) is proposed based on YOLOv3 in this study. It is the first time to represent a WSI via feature vectors corresponding to $32\times32$ small areas. In this new framework, effectiveness and efficiency are both improved. And, the fully convolutional structure of YOLCO with a light weight can avoid the same computation by extremely increasing the input size.

For encoding WSI by feature vectors, feature mixture are commonly and naturally established by the multi-task supervision for the CNN in one mixed loss function. Dual tasks, classification and location, in this study have a special function for the cytological data while the natural data does not have. That is the location annotation is more stable than the semantic based on the physical property.
%Traditional approaches for controlling the attention of the model for each task use weighted losses for different tasks, for instance, which is valid as well as focal loss to deal with unbalanced data. For sparse cervical cells in the patch-level image, the focal loss is adapted to improve the performance of our single step CNN.

%As highlighted by colors in Table \ref{is-ablation},
It is reasonable that feature vectors of YOLCO with both semantic and location information are robust. The network for single task with a larger number of parameters can be well fitted at the patch-level just like upstream studies. Because the large amount of unseen data at the WSI-level, however, it will challenge the generalization of the model (the problem of the overfitting). So, when testing all WSIs, the trend of performance becomes opposite to that of patch-level. While the proposed YOLCO weakened the above effects because of the use of feature mixture by both classification and location tasks in this downstream study, and always performs the best in different settings of the WSI dataset.

Otherwise, YOLCO is weak for the natural data. The comparison in the public MS COCO 2014, which has 40775 training images into 80 classes, is inconsistent with cytopathology:
{\footnotesize\centering\begin{tabular}{p{1.775cm}p{1.775cm}p{1.775cm}p{1.775cm}}
 & Tiny & YOLOv3 & YOLCO \\\hline
 mAP$_{.5}$  & 0.348 & 0.579 & 0.277 \\
\end{tabular}}
Dissimilar with the performance in the cervical data, the proposed YOLCO has a meagre performance in the natural dataset, $27.7\%$ at mAP. It is not effective in such a complex upstream task because there are 80 classes and wider texture variety than the cytopathology. Not to mention the difficulty in overlapped objects and the depth of field, which are rare and slight in cytopathology.

\section{Conclusion}

In conclusion, the proposed YOLCO using the novel InCNet module achieves a better WSI-level performance than the conventional framework and the SOTA model, while a slight loss occurs to the detection at the patch-level. Remarkably, the speed of YOLCO is faster and lighter than the SOTA CNNs and, meanwhile, the accuracy is better, which proves the workability and the efficiency of the proposed method via the simplified manner and the valid InCNet.

The experimental results further proved that well-designed models for the natural upstream task are improvable for the particular downstream task. Especially, the multi-connection between the different sizes of the receptive field inner the module can enrich the feature representation without any residual-like connections to deal with various complexities of cervical cells, while common CNNs prone to equally encode every pattern of the cervical cell image is inferior for the WSI-level analysis of cytopathology. It is reliable that the performance on the cervical data will be improved when the structure of the model can be further optimized according to our experiments. Moreover, the limited local annotation at the patch-level is relatively acceptable for multi-cohort and multiscanner slides based on our feature mixture from the multiple task learning and the enhancement of the multi-scale inline connection.

\section*{Acknowledgments}
The authors want to thank pathologists and organizations that provided the raw data and the manual annotations. As well as the Collaborative Innovation Center for Biomedical Engineering and the Britton Chance Center and MOE Key Laboratory for Biomedical Photonics should be greatly appreciated for their platforms and devices. This work is supported by the NSFC projects (grant 61721092) and the director fund of the WNLO.

%%Harvard
\bibliographystyle{model2-names.bst}\biboptions{authoryear}
\bibliography{refs}
%
%\section*{Supplementary Material}
%
%Supplementary material that may be helpful in the review process should
%be prepared and provided as a separate electronic file. That file can
%then be transformed into PDF format and submitted along with the
%manuscript and graphic files to the appropriate editorial office.

\end{document}